\documentclass[10pt,twocolumn,letterpaper]{article}
\usepackage{booktabs}
\usepackage{siunitx}
\usepackage{tikz}
\usepackage{pgfplots}
\usepackage{multirow}
\pgfplotsset{compat=1.18}
\usepackage{wacv}              

\usepackage{graphicx}
\usepackage{amsmath}
\usepackage{amssymb}
\usepackage{booktabs}
\usepackage{tablefootnote}
\usepackage[accsupp]{axessibility}

\newcommand{\acc}{Acc$_{EF}$}
\newcommand{\ourmodel}{EchoDFKD}
\newcommand{\ourmodelone}{EchoDFKD$_1$}
\newcommand{\ourmodeltwo}{EchoDFKD$_2$}
\newcommand{\ourmodelthree}{EchoDFKD$_3$}
\newcommand{\ourmodelfour}{EchoDFKD$_4$}
\newcommand{\aFDED}{aFD$_{ED}$}
\newcommand{\aFDES}{aFD$_{ES}$}
\newcommand{\meanIoU}{\text{meanIoU}}
\definecolor{bluefig}{HTML}{6C92C1}
\definecolor{orangefig}{HTML}{D1B864}
\definecolor{redfig}{HTML}{666666}
\usepackage[pagebackref,breaklinks,colorlinks]{hyperref}

\usepackage[capitalize]{cleveref}

\crefname{section}{Sec.}{Secs.}
\Crefname{section}{Section}{Sections}
\Crefname{table}{Table}{Tables}
\crefname{table}{Tab.}{Tabs.}

\def\wacvPaperID{636} 
\def\confName{WACV}
\def\confYear{2025}

\begin{document}
\title{EchoDFKD: Data-Free Knowledge Distillation for Cardiac Ultrasound Segmentation using Synthetic Data}

\author{
Grégoire Petit\textsuperscript{1,2,*}, 
Nathan Palluau\textsuperscript{*}, 
Axel Bauer\textsuperscript{2},
Clemens Dlaska\textsuperscript{1,2}\\ 
 \textsuperscript{1}Digital Cardiology Lab, Medical University of Innsbruck, A-6020 Innsbruck, Austria\\
 \textsuperscript{2}University Clinic of Internal Medicine III, Cardiology and Angiology,\\Medical University of Innsbruck, A-6020 Innsbruck, Austria\\
{\tt\small g.petit360@gmail.com},~{\tt\small nathan.palluau@gmail.com},~{\tt\small clemens.dlaska@i-med.ac.at}
}

\maketitle


\begin{abstract}
The application of machine learning to medical ultrasound videos of the heart, i.e., echocardiography, has recently gained traction with the availability of large public datasets. Traditional supervised tasks, such as ejection fraction regression, are now making way for approaches focusing more on the latent structure of data distributions, as well as generative methods.

We propose a model trained exclusively by knowledge distillation, either on real or synthetical data, involving retrieving masks suggested by a teacher model. We achieve state-of-the-art (SOTA) values on the task of identifying end-diastolic and end-systolic frames. By training the model only on synthetic data, it reaches segmentation capabilities close to the performance when trained on real data with a significantly reduced number of weights.
A comparison with the 5 main existing methods shows that our method outperforms the others in most cases.

We also present a new evaluation method that does not require human annotation and instead relies on a large auxiliary model. We show that this method produces scores consistent with those obtained from human annotations. Relying on the integrated knowledge from a vast amount of records, this method overcomes certain inherent limitations of human annotator labeling.
\end{abstract}~\\
Code:~\href{https://github.com/GregoirePetit/EchoDFKD}{https://github.com/GregoirePetit/EchoDFKD}

\begin{figure*}
    \centering
\includegraphics[width=\linewidth]{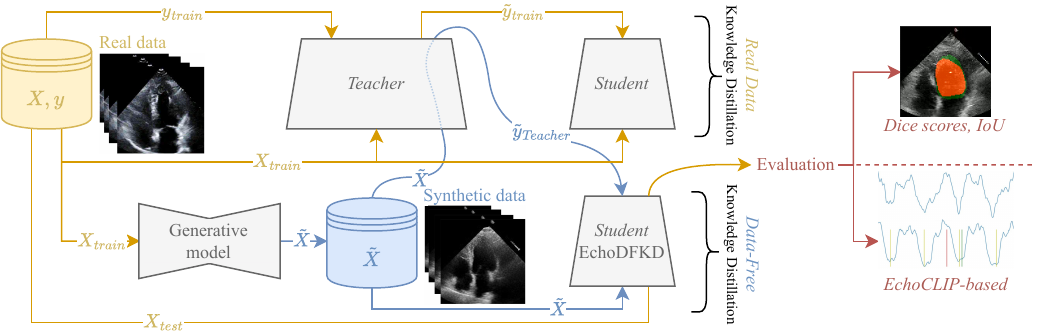}
    \caption{Overview of \ourmodel. In Knowledge Distillation, \textcolor{orangefig}{real data}, e.g., from EchoNet-Dynamic~\cite{ouyang2020echonet}, is often used to train a \textcolor{redfig}{\textit{Teacher}} model and then used to train a \textcolor{redfig}{\textit{Student}} model. \ourmodel~is using \textcolor{bluefig}{synthetic data} from EchoNet-Synthetic dataset~\cite{reynaud2024EchoNetsynthetic} to distill knowledge. By analyzing the mask generated by our ConvLSTM-based \ourmodel~segmentator or the similarity outputs of custom EchoCLIP~\cite{christensen2024echoclip} prompts and EchoNet-Dynamic raw images, we can predict the average Frame Distance (aFD). Additionally, our~\ourmodel~segmentator described in Subsec.~\ref{subsec:archi}, is evaluated against EchoCLIP knowledge to assess its segmentation quality alongside traditional metrics (\meanIoU, dice score) against human labels.\vspace{-12pt}}

    \label{fig:overview}
\end{figure*}
\def\thefootnote{*}\footnotetext{equal contribution}

\def\thefootnote{\arabic{footnote}}
\section{Introduction}
\label{sec:intro}
In Deep Learning applied to computer vision, one critical task is segmentation \cite{ronneberger2015u}, which is essential for accurately interpreting and analyzing visual data. However, segmentation as an annotation task is often resource-intensive and time-consuming, particularly in the medical domain. Knowledge distillation (KD) ~\cite{hinton2015distilling}  typically aims to transfer knowledge, such as the ability to segment data, from a large, complex model (often referred to as the ``teacher model") to a smaller, more efficient model (the ``student model"). This process involves training the student model to replicate the outputs of the teacher model on certain tasks. By doing so, the student model inherits the teacher's capabilities relevant to this task, requiring fewer computational resources. Beyond model compression, KD has demonstrated its versatility in various applications, such as adversarial robustness~\cite{papernot2016distillation}, ensemble fusion~\cite{you2017learning}, continual learning~\cite{szatkowski2024adapt}, partial or missing labels~\cite{Camporese2020action}, multi-task learning~\cite{li2020knowledge}, and cross-modal learning ~\cite{aytar2016soundnet}.

It often happens that the medical data used to optimize a model is not accessible while the model itself is shared. In particular, regarding computer vision applied to echocardiography, several recent models are shared with the public while the corresponding datasets are not~\cite{zhang2018fully,christensen2024echoclip}.
It's a safe bet that many laboratories are reluctant even to share their models because of the risk of reconstruction attacks. Knowledge distillation offers protection against attacks on sensitive biometric data that attempt to reconstruct training examples from models, making it a robust solution in medical applications~\cite{papernot2016semi}. Owners of a sensitive dataset could, therefore, share only the outputs of their models on several examples or a fusion of outputs from restricted models instead of sharing their model directly. Furthermore, KD is an efficient way of transferring knowledge from one modality to another and thus managing the situation common in medicine, where the different modalities of an example are not always the same from one dataset to another (some datasets might include, for instance, other views or patient metadata). Finally, it is effective for model compression when the student model is lightweight, reducing noise and forcing the student model to focus on the knowledge of interest while removing, for instance, labeler-specific style.

Compared with KD methods based on real data, Data-Free Knowledge Distillation (DFKD) methods enable knowledge to be distilled on a potentially infinite number of artificial examples, making it possible to achieve a near-perfect imitation of the teacher model, at least in the domain well represented by the generative model~\cite{lopes2017data,chen2019data,yoo2019knownledgeextraction}. In addition, it is possible to focus on a particular section of the possible examples space, e.g., by conditioning on a particular part of a patient's health data (such as age, resting heart rate, etc.), thus generating a very large number of examples under these conditions to ensure efficient distillation of one or more teachers to a specialized student. Moreover, DFKD methods are not necessarily incompatible with knowledge distillation methods based on real data since synthetic data can be used in addition to real data, in pre-training, in composite batches, or even in the form of data augmentation when examples are created from real examples. \cite{azizi2023synthetic,he2023synthetic}.

In the present work, we introduce EchoDFKD, the first DFKD model for echocardiography. We use this paradigm by relying on the EchoNet-Synthetic dataset~\cite{reynaud2024EchoNetsynthetic}. This dataset was generated following the example of EchoNet-Dynamic (see Figure~\ref{fig:overview} and Section~\ref{sec:method}), whose test set we use to demonstrate the suitability of our approaches for subsequent tasks on real data.

We first provide the performance of our models using traditional methods based on human labels to provide a fair comparison with models achieving SOTA performance and show that the very lightweight convolutional long short-term memory (ConvLSTM) architecture~\cite{shi2015convolutional,shi2017deep}, as some previous results using this architecture (but with more weights) had suggested ~\cite{li2019recurrent}, are relevant for  apical-4-chamber echocardiograms. Furthermore, we establish a line on the Pareto (score, weight) frontier (see Figure~\ref{fig:loglog}), which demonstrates that, although the need to process frames one by one requires relatively large numbers of floating point operations per second (FLOPS)~\cite{maani2024simlvseg}, it is possible to achieve competitive performance levels utilizing a significantly lower number of FLOPS.

To take our model-centric approach to its full extent, we propose removing humans from the process altogether by proposing evaluation methods by other models. This demonstrates that starting from initial models that have been developed involving human knowledge, it is then possible to successfully tackle machine learning tasks by letting these models interact with each other. In the context of unannotated medical data availability, this offers the advantage of enabling rigorous evaluation of masks provided by another agent on this data and fair comparison with other mask proposals. Indeed, while model-based evaluation is highly error-prone, these errors are less likely to be systematic or style-specific. 

\section{Related Work}
\label{sec:sota}
\subsection{Data-Free Knowledge Distillation}
Data-Free Knowledge Distillation is a technique in which a student model is trained to replicate the behavior of a teacher model without access to the original training data. 
An alternative twisted approach involves aggregating outputs to train a student model and sharing only the student model, as seen in PATE~\cite{papernot2016semi}, which discusses medical data. Another method creates virtual images from the teacher model, using `metadata' (e.g., means and standard deviations of activations from each layer) recorded from the original training dataset. However, these metadata are often not provided for well-trained CNNs~\cite{lopes2017data}.

A standard approach utilized in DFKD involves using a GAN (including multiple variants) and distilling a teacher model. This approach updates the generator based on the produced distribution at each step, extracts the penultimate layer from the teacher, and then evaluates it~\cite{chen2019data}. Similarly, KegNet~\cite{yoo2019knownledgeextraction} focuses on extracting trained deep neural network knowledge and generating artificial data points to replace missing training data in knowledge distillation.

\subsection{Large Vision Models evaluation}
Vision Transformers are known to be effective mask auto-labelers~\cite{lan2023vision}, which suggests that using pseudo-labels to evaluate models is not unreasonable. Additionally, some approaches aim to find a latent ground truth among multiple annotators, acknowledging that disagreement among annotators is not necessarily problematic~\cite{guan2018said}. Furthermore, it is also common to use Large Language Models (LLMs) for labeling tasks~\cite{tan2024large}.

In our work, we examine the quality of Large Vision Models (LVMs) labeling. Since our segmentations do not originate from LVMs, we can use them to evaluate our segmentations bias-freely.

\subsection{Segmentation}

The field of ultrasound image segmentation using machine learning methods is currently very active.
Efforts in fully automated echocardiogram interpretation are highlighted in~\cite{zhang2018fully}, which, although trained on a private data set, laid the foundations for future research. The introduction of the CAMUS dataset, as detailed in~\cite{leclerc2019deep}, provided a well-labeled but smaller dataset for segmentation tasks.

A significant milestone was the release of the EchoNet-Dynamic dataset~\cite{ouyang2020echonet}, which has become a benchmark for many studies~\cite{rajpurkar2022ai,shamshad2023transformers,ma2024segment,bommasani2021opportunities}. The segmentation and prediction of ejection fraction (EF) using this dataset has been explored extensively~\cite{kazemi2020deep}.

Recent advancements have seen the application of transformer models and other novel architectures. To predict end-systolic (ES) and end-diastolic (ED) frames, \cite{Reynaud2021aFD} employed a BERT model by treating the sequence like a series of words. Similarly, \cite{fazry2022hierarchical, muhtaseb2022echocotr} focused on both segmentation and EF prediction, the later achieving SOTA results on EF predictions.

Methods using pre-training~\cite{saeed2022contrastive} seem to be particularly effective. Notably, \cite{maani2024simlvseg}, from the same laboratory as EchoCoTr, achieves the best segmentation results to date.

Innovations in lightweight and explainable models have also emerged. For EF estimation, \cite{mokhtari2022echognn} used graph neural networks with a focus on ED and ES frames, while~\cite{muldoon2023lightweight} achieved commendable segmentation and EF prediction using a lightweight Mobile U-Net~\cite{ronneberger2015u}.

\section{Proposed Method}
\label{sec:method}
The overview of the proposed method, illustrated in Figure~\ref{fig:overview} is detailed in the following sections.
\subsection{Datasets}
The release of EchoNet-Dynamic dataset~\cite{ouyang2020echonet} by the Stanford University Medical Center, which contains 10,024 sparsely annotated\footnote{Six additional videos can be found in the original publicly available dataset, but they are not annotated. Examination of the code from the original paper and the codes shared in subsequent research works on the dataset shows that they are always filtered at the beginning of the pipeline, meaning that, in practice, only 10,024 clips are processed.} using a classic apical-4-chamber view image tracing method, has stimulated deep learning-oriented research on the interpretation of echocardiography videos.
Annotations would consist of approximating the segmentation mask of the left ventricle using segments. Generally, one segment is along the principal axis, and then twenty segments cross this axis. Some examples have been annotated multiple times. The dataset contains examples with varied sampling rates and different image qualities. Some clips are corrupted (see supplementary material).

We also use the EchoNet-Synthetic dataset~\cite{reynaud2024EchoNetsynthetic}, which provides the same data type but is generated by the recent XSCM generative model~\cite{reynaud2023feature}. This external, synthetic dataset enables us to train data-free. Because of its generative nature, the video clips are not that close to the original EchoNet-Dynamic dataset on which it was trained. However, the displacement fields across the samples are good enough to represent the heart motion's basics accurately. Furthermore, as the Ejection Fraction conditions the generative network, the synthetic distribution ultimately becomes close enough for our KD setup.

\subsection{Architecture}
\label{subsec:archi}
The model architecture we use extends the classic U-Net~\cite{ronneberger2015u} design by integrating ConvLSTM~\cite{shi2015convolutional} layers, forming a hybrid structure particularly well suited for spatiotemporal processing. The U-Net component handles spatial feature extraction through its characteristic encoder-decoder structure, utilizing convolutional layers for downsampling and upsampling, thereby preserving fine-grained details essential for segmentation. The ConvLSTM layers are embedded within the U-Net to capture temporal dependencies across sequences of echocardiographic images. We systematically investigated configurations ranging from a single block of one ConvLSTM layer to four blocks of four ConvLSTM layers, aiming to determine the optimal balance between model complexity and performance. We denote by (B$b$,l$l$) the \ourmodel~that the downsampling part comprises $b$ blocks of $l$ ConvLSTM layers each. By maintaining a hidden state that evolves over time, these layers enable the model to learn temporal patterns and transitions between consecutive frames, which is critical for accurate segmentation in sequences. This hybrid architecture allows the model to effectively map input sequences of images to output sequences of segmentation masks, enhancing its ability to perform tasks where both spatial and temporal information are crucial, such as in the segmentation of cardiac cycles.

\subsection{Lightweight Models for Specific Tasks}

Using the lightest possible model capable of performing the specific assigned task is advantageous for typical healthy cases. Heavier models represent the function $f:record\rightarrow verdict$ in a high-dimensional space, which can lead to two main issues: (1) Overfitting to Specific Annotation Styles: This can result in the model adapting to systematic human errors or institution-specific recording peculiarities, as seen in current models whose performance drops significantly across different datasets. Many medical datasets, including EchoNet-Dynamic, compile several composite sources with recognizable formats (due to cropping methods, sampling rates, etc.). Each source may be associated with particular annotators. A heavy model tends to learn links between input specificities and annotation specificities, which can be mitigated during compression. (2) Focusing Excessively on Outlier or Corrupted Examples: Heavier models may disproportionately emphasize particular, often corrupted examples that may not strictly pertain to the defined task. 

It is important to note that the models we use here never exceed 4 million parameters.

\subsection{Large Vision Models (LVMs) and Specialized Models Evaluation}

The emergence of large models like CLIP~\cite{radford2021clip} and some specialized versions of it like EchoCLIP~\cite{christensen2024echoclip}, capable of performing multiple tasks, offers the potential for generating pseudo-labels that complement human labels. These large models can be viewed as valuable agents in producing opinions. Their outputs, obtained at a lower cost than labor-intensive human segmentation labels, can be used to curate training datasets or refine evaluation sets by distinguishing between examples that fit a defined framework and those that do not.

Large models can also assess the understanding of specialized smaller models regarding the signals they are supposed to master. For example, they can verify that these smaller models can retrieve task-related metrics. This approach helps avoid issues related to noise and biases inherent in human labels, particularly those requiring high precision and susceptible to annotator fatigue. Additionally, imperfect pseudo-labels can integrate basic knowledge from private learning and multimodal practitioner analysis, such as annotations made with access to multiple perspectives.

\subsection{Practical Considerations in Assessing and Comparing Models}

A model should be evaluated based on the specific task it is designed to perform. It is unrealistic to expect a model to handle all out-of-distribution cases effectively. In medical computer vision applications, practitioners generally prefer a model that recognizes its limitations and alerts the user when the input signal is unusual rather than a model trained to provide mediocre results in poorly captured cases.

It is important to remember that clips used for machine learning are often short extracts from longer signals. A user can extend the recording if a model indicates it cannot interpret the signal in real-time. This justifies the transition to human examination for patients with anomalies that cause deviation from the usual distribution. Thus, these examples can be considered positives in the context of pre-screening with the model.

This approach ensures the model remains reliable and useful in real-time applications, distinguishing between typical cases and those requiring further human intervention.

\subsection{Data-Free Knowledge Distillation Setting}
\noindent\textbf{Synthetic Data Generation~~~} The student model is trained on synthetically generated data crafted to resemble the distribution of the original training data used for the teacher model.\\
\noindent\textbf{Knowledge Transfer~~~} The teacher model provides supervision by generating pseudo-labels or masks for the synthetic data. The student model then learns to replicate these outputs, effectively distilling the teacher's knowledge without direct access to real data.\\
\noindent\textbf{Robustness Assessment~~~} Additional evaluations may include robustness tests to ensure the student model performs well across diverse and unseen data subsets and highlights its generalization capabilities.
\subsection{Training}

We conducted a limited exploration of hyperparameters without delving too deeply. Among the hyperparameters, we included choices between different parameter initialization methods, various depths (number of channels) in intermediate layers, different losses, different batch sizes, different sequence lengths, and various learning rate management strategies. For models with multiple blocks, we also allowed ConvLSTM to create residual connections between the outputs of different blocks. Notably, hyperparameter optimization selected a residual connection for the last block in our best-performing model.

As the teacher model, we chose the model shared when EchoNet-Dynamic was released, based on a DeepLabv3 architecture \cite{ouyang2020echonet}.

\begin{figure}[h]
    \centering
    \resizebox{\linewidth}{!}{
    \begin{subfigure}{0.15\textwidth}
        \includegraphics[width=\linewidth]{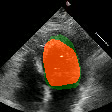}
        \caption*{(a)}
    \end{subfigure}
    \hfill
    \begin{subfigure}{0.15\textwidth}
        \includegraphics[width=\linewidth]{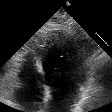}
        \caption*{(b)}
    \end{subfigure}
    \hfill
    \begin{subfigure}{0.15\textwidth}
        \includegraphics[width=\linewidth]{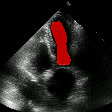}
        \caption*{(c)}
    \end{subfigure}
    }\vspace{-5pt}
    \caption{Segmentation examples from an EchoNet-Dynamic video [(a), (b)] and an EchoNet-Synthetic video [(c)]. In (a), EchoDFKD predictions are shown in the green (G) channel, while DeepLabv3 predictions are displayed in the red (R) channel. Image (b) is the original, unaltered EchoNet-Dynamic frame. Image (c) represents a segmentation mask of DeepLabv3 on the EchoNet-Synthetic dataset used to train \ourmodel.}\vspace{-10pt}
    \label{fig:exemple}
\end{figure}
The models converged in approximately 400 epochs. We selected the model from the epoch that yielded the best validation loss. The validation set was used for hyperparameter optimization and for choosing the thresholds to convert floating-point masks to boolean masks.

\subsection{Direct Evaluation of Segmentation: Assessing mask quality} 

Creating a segmentation label is more complex than producing labels for tasks like classification or regression. This complexity arises because a segmentation mask requires assigning a label to every pixel in an image, effectively working in a much higher-dimensional space. Unlike classification or regression, where a single decision is made for the entire image, segmentation demands precise labeling of intricate details across the whole picture. Due to this complexity, even expert human annotators often find it challenging to produce perfectly accurate segmentation masks, which can lead to inconsistencies and a lack of satisfaction with the labels they create.

For instance, if we refer to the value of the left ventricular ejection fraction (LVEF), which is often derived from multiple segmentation masks, it is interesting to note that the best performance in estimating the LVEF on EchoNet-Dynamic, currently achieved by EchoCoTr~\cite{muhtaseb2022echocotr} (the coefficient of determination \( R^2 \) is 0.82, meaning that the squared correlation coefficient between outputs and targets is at least 0.82), surpasses the intra-annotator variability (which squared correlation coefficient is \( r^2 \) = 0.81) reported in~\cite{leclerc2019deep}. Although these values are not truly comparable, as the annotator named O1 from CAMUS does not necessarily have the same consistency as that of DeepLabv3, and EchoNet-Dynamic is different from that of CAMUS, which also includes 2-chamber views, this comparison still raises methodological questions. It should be noted that the intra-annotator correlation coefficient cannot in any way constitute a theoretical ceiling for the performance of a model; the model outputs could be extremely close to the mean annotation of the annotator while being less noisy, the proof can be found in the supplementary materials. However, cases of exceedances warrant attention.

In EchoNet-Dynamic, the human labels consist of 21 segments, which cannot be strictly classified as segmentation but as a polygonal annotation. It is easy to understand that a label produced this way cannot be perfectly satisfactory, as some regions are not labeled at the pixel level. When EchoNet-Dynamic was released, the authors showed the segmentations produced by their model and the segmentations produced by humans, and some of the segmentations produced by the machine were preferred to that of the human. This approach can be likened to second-order labeling, like that used to train a reward model.

Inspired by this observation, we chose to use the large EchoCLIP~\cite{christensen2024echoclip} model to assess the quality of the masks produced by the models. Large vision models trained by constrastive learning have already demonstrated their ability to perform segmenting tasks~\cite{fan2020learning,yao2021filip,yi2023simple}. 
Without going that far, we are simply trying to use such a model to judge the quality of semantic segmentation.

We rely on EchoCLIP's ability to judge the quality of a segmentation mask based on a few well-chosen prompts. To do this, we adopt a strategy that evaluates, on the one hand, that the mask does not significantly exceed the boundaries of the left ventricle and, on the other hand, that it adequately fills the entire area of the left ventricle.

To evaluate whether the mask does not overflow, we check that EchoCLIP can correctly distinguish the walls around the left ventricle. For this purpose, we blacken the area of the image under the mask and show the frames to EchoCLIP, using the prompt ``\textit{WALL}" to verify that it correctly distinguishes the walls of the left ventricle. 
\begin{figure}[h!]
    \centering\vspace{-8pt}
    \resizebox{\linewidth}{!}{\begin{tikzpicture}
    \pgfplotsset{
        every axis x label/.style={at={(ticklabel cs:0.5)},anchor=north,yshift=-8pt},
    }

    \begin{axis}[
        width=7.55cm, height=5cm,
        xlabel={Number of parameters of our model},
        xlabel style={yshift=12pt},
        xtick=data,
        xticklabels={6K, 16K, 27K, 37K, 40K, 92K, 144K, 196K, 175K, 393K, 612K, 830K, 715K, 1.6M, 2.5M, 3.4M},
        x tick label style={rotate=59, anchor=east},
        ymin=0.61, ymax=0.95,
        xmin=1, xmax=16,
        ylabel={\textcolor{redfig}{meanIoU}, \textcolor{bluefig}{Dice score}},
        y label style={yshift=-8pt},
        axis y line*=left,
        ytick style={color=black},
        ylabel near ticks,
    ]

    \addplot[redfig, thick] table[row sep=\\]{
        1 0.6297 \\
        2 0.6491 \\
        3 0.7294 \\
        4 0.7458 \\
        5 0.7936 \\
        6 0.7908 \\
        7 0.8036 \\
        8 0.7986 \\
        9 0.8136 \\
        10 0.8193 \\
        11 0.8201 \\
        12 0.8252 \\
        13 0.8250 \\
        14 0.8277 \\
        15 0.8271 \\
        16 0.8269 \\
    };

    \addplot[bluefig, thick] table[row sep=\\]{
        1 0.7531 \\
        2 0.7684 \\
        3 0.8303 \\
        4 0.8438 \\
        5 0.8779 \\
        6 0.8786 \\
        7 0.8880 \\
        8 0.8847 \\
        9 0.8944 \\
        10 0.8986 \\
        11 0.8991 \\
        12 0.9023 \\
        13 0.9021 \\
        14 0.9039 \\
        15 0.9037 \\
        16 0.9034 \\
    };

    \end{axis}

    \begin{axis}[
        width=7.55cm, height=5cm,
        ymin=0.0072, ymax=0.0091,
        xmin=1, xmax=16,
        ylabel={\textcolor{orangefig}{EchoCLIP score}},
        axis y line*=right,
        ytick style={color=black},
        ylabel near ticks,
        yticklabel style={anchor=west},
        axis x line=none,
    ]

    \addplot[orangefig, thick] table[row sep=\\]{
        1  0.007744114 \\
        2  0.007789685 \\
        3  0.008201673 \\
        4  0.008172777 \\
        5  0.0085004885 \\
        6  0.008498515 \\
        7  0.008612661 \\
        8  0.008559526 \\
        9  0.008552759 \\
        10 0.008620328 \\
        11 0.0086194575 \\
        12 0.008674325 \\
        13 0.0086746635 \\
        14 0.008622474 \\
        15 0.008593912 \\
        16 0.008670824 \\
    };

    \end{axis}

\end{tikzpicture}}\vspace{-5pt}
    \caption{Illustration of the relationship between the number of model parameters and three key performance metrics: mean Intersection over Union (meanIoU), Dice score, and our custom EchoCLIP score. The meanIoU and Dice score, displayed on the left y-axis, show how segmentation accuracy against human annotators improves with increased model complexity. Our EchoCLIP score, shown on the right y-axis, reflects the segmentation quality without needing any annotator. In the EchoCLIP segmentation quality assessment, the segmentation quality is determined by the difference between the prompts ``\textit{LEFT VENTRICLE}" and ``\textit{NOTHING}" applied on raw masks that have been expanded by a few pixels.}
    \label{fig:fig34}
\end{figure}
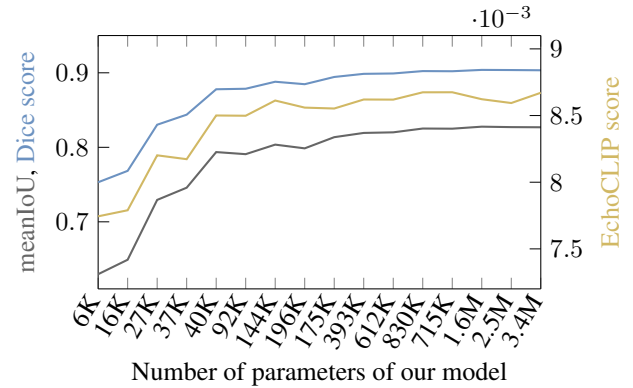
To evaluate whether the mask is not too small, we take the opposite approach; we enlarge the mask by a few (5) pixels at the borders, then blacken the enlarged mask. We check that EchoCLIP recognizes the resulting image by differentiating the responses to the prompts ``\textit{LEFT VENTRICLE}" and ``\textit{NOTHING}". Figure~\ref{fig:fig34} represents the evolution of EchoCLIP answers depending on the model size. We observe a plateau beyond which the model no longer improves, confirmed by the evolution of the dice scores and meanIoU, detailed in Table~\ref{tab:metrics}, measured against the frames initially labeled in EchoNet-Dynamic.

\begin{table}[t]
\centering
\resizebox{\linewidth}{!}{
\begin{tabular}{rrcccc}
\hline
  && B1 & B2 & B3 & B4 \\ \toprule
\multirow{4}{*}{meanIoU}    & l1 & 66.92\% & 77.31\% & 82.47\% & 83.70\% \\ 
                            & l2 & 66.23\% & 81.08\% & 83.30\% & 83.96\% \\ 
                            & l3 & 73.38\% & 81.19\% & 83.53\% & 83.89\% \\ 
                            & l4 & 74.47\% & 81.59\% & 83.62\% & 83.72\% \\ \midrule
\multirow{4}{*}{Dice score} & l1 & 78.92\% & 86.50\% & 90.13\% & 90.93\% \\ 
                            & l2 & 78.21\% & 89.17\% & 90.68\% & 91.11\% \\ 
                            & l3 & 83.68\% & 89.23\% & 90.83\% & 91.07\% \\ 
                            & l4 & 84.48\% & 89.58\% & 90.89\% & 90.96\% \\ 
 \midrule
 \end{tabular}}
 \resizebox{\linewidth}{!}{

\begin{tabular}{rll}
 DeepLabv3~\cite{ouyang2020echonet}~~~~~~~~~~~~~~~~ & Dice score~~~~~~~~~~~~~~~ & 92.26\% \\
 simLVSeg~\cite{maani2024simlvseg}~~~~~~~~~~~~~~~~  & Dice score & 93.32\% \\
 MU-UNET~\cite{muldoon2023lightweight}~~~~~~~~~~~~~~~~ & Dice score & 90.50\% \\
 nnU-net\tablefootnote{reported from~\cite{maani2024simlvseg}}~\cite{isensee2021nnu}~~~~~~~~~~~~~~~~ & Dice score & 92.86\% \\
 \bottomrule
\end{tabular}
}
\caption{Traditional performance metrics across \ourmodel~configurations (B$b$,l$l$)\tablefootnote{described in Subsection~\ref{subsec:archi}}, against human annotators. DeepLabv3~\cite{ouyang2020echonet}, simLVSeg~\cite{maani2024simlvseg}, MU-UNET~\cite{muldoon2023lightweight} and nnU-net~\cite{isensee2021nnu} are trained on real data, whereas \ourmodel~is trained via knowledge distillation on synthetic data.
\vspace{-10pt}}\label{tab:metrics}
\end{table}

\subsection{Evaluation on a Downstream Task } 

Because judging a model based on the segmentation task is difficult, as human labels are imperfect in their own opinion, it is natural to judge the knowledge acquired by the model based on its ability to perform a task whose completion seems possible for an agent who knows how to perform the main task, such as reconstructing features that can be deduced from the segmentation.
In particular, the ability to identify the end-diastolic (ED) and end-systolic (ES) frames constitutes a robust test, which has already been employed on other datasets \cite{dezaki2017deep}, and seems to have recently attracted attention for the EchoNet-Dynamic dataset ~\cite{Reynaud2021aFD,mokhtari2022echognn}.
These frames have a concrete utility as they allow the calculation of LVEF when segmented.

We noticed that the literature's method for determining aFD tends to be inconsistent. 
It can be calculated using sequences of different lengths and different strategies. For example, in~\cite{mokhtari2022echognn}, sub-clips of 64 frames are used, and results are only provided for examples where different phases are identified. They discard from the evaluation examples where their model does not find one of the two phases. In~\cite{Reynaud2021aFD} sub-clips of size 128 are extracted using a method that is difficult to assess in terms of statistical artifacts and also remove some examples that the model is unable to process (between 3 and 6 in the case of the scores reported here). We, therefore, propose a simple, direct method for deducing these frames from any model that can produce, in zero-shot, a sequence of values (one per frame) correlated with the alternation of the two phases and hope that it can constitute a standard. In addition, we advise not to remove any more examples during evaluation, to facilitate comparison between models. We report our performance both on the entire test set (we take these values to compare ourselves with other works) and, for information purposes only, on a subset that excludes examples with corrupted labels.\vspace{-10pt}

\paragraph{Methodology}
(1)~We take the entire video clip,
(2)~We examine the masks produced by the model and calculate the mask area in each frame. We calculate the median of the resulting sequence.
(3)~We identify the contiguous blocks of values below (or above) the median (this roughly corresponds to systolic or diastolic phases), choose the contiguous block closest to the reference value (to position ourselves on the same beat), and select the smallest (or highest) value within this block from the series of areas.

\begin{figure}
    \centering
    \resizebox{0.45\linewidth}{!}{\begin{tikzpicture}
    \begin{axis}[
        xlabel={Sampling rate},
        ylabel={Mean aFD error},
        grid=both,
        grid style={dashed, gray!30},
        major grid style={line width=0.2pt, draw=gray!50},
        minor grid style={line width=0.1pt, draw=gray!30},
        width=4.5cm,
        height=4.5cm,
        legend style={at={(0.05,0.95)}, anchor=north west},  
        ]

    \addplot[
        only marks,
        color=bluefig,
    ] table [x=x, y=y, col sep=comma] {figures/tikz/mean_aFD_err/plot-data.csv};

    \addplot[
        red,
        thick,
        domain=20:120,  
    ]
    {0.0531097888*x + 0.08233129047209967};

    \end{axis}
\end{tikzpicture}}\vspace{-8pt}
    \caption{Relationship between sampling rate and mean aFD error for \ourmodel: This plot shows the mean aFD (average Frame Distance) error of the EchoNet-Dynamic dataset as a function of the sampling rate used for EchoDFKD.
\vspace{-10pt}}
    \label{fig:erroraFD}
\end{figure}
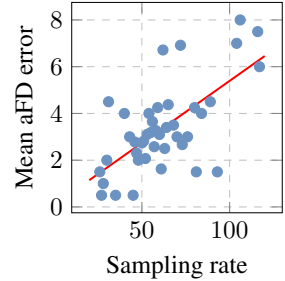

Note that, as might be expected, the errors in the aFD computing, shown in Figure~\ref{fig:erroraFD}, appear to be proportional to the sampling rate. Consequently, comparing aFD scores from one dataset to another is a delicate matter, as this requires a certain homogeneity in the sampling rate distribution.\vspace{-10pt}

\paragraph{Alternative way of evaluating performances}

Comparing the frame labeled as a systole by the human annotator to the corresponding frame automatically labeled as a systole based on the mask widths from the original EchoNet-Dynamic's DeepLabv3, we observe that the human-identified frame often appears before the one identified by DeepLabv3, on average, 2.3 frames earlier.

Our interest in finding an unbiased ground truth motivates our approach to using EchoCLIP to discriminate between ED and ES frames and other frames. To do this, we attempted to find prompts that varied simultaneously with the size of the DeepLabv3 masks and could capture this quasi-cyclical movement of the image. 

\begin{figure}[h]
    \centering
    \resizebox{\linewidth}{!}{\begin{tikzpicture}[scale=0.8]
    \begin{axis}[
        title={\parbox{.5\textwidth}{\centering Performance of EchoCLIP prompts on 0X13ACF1360A23F50F}},
        xlabel={Frame number},
        ylabel={Normalized values},
        legend cell align={left}, 
        major grid style={line width=0.2pt, draw=gray!50},
        minor grid style={line width=0.1pt, draw=gray!20},
        ymin=-2, ymax=3.1,
        xmin=0, xmax=160,
        width=10cm,
        height=5cm
    ]

    \addplot[
        color=red,
        line width=1pt
    ] table [x=x, y=y, col sep=comma] {figures/tikz/comparison/DeepLabv3.csv};
    \addlegendentry{DeepLabv3 Mask Area}

    \addplot[
        color=bluefig,
        line width=1pt
    ] table [x=x, y=y, col sep=comma] {figures/tikz/comparison/cumsum.csv};
    \addlegendentry{\textit{f}}
    \end{axis}
\end{tikzpicture}} 
    \caption{Comparison of DeepLabv3 masks area and prompts similarities. \textit{f} is the cumulative sum of the difference between the prompts ``\textit{THE MITRAL VALVE IS CLOSED}" and ``\textit{THE MITRAL VALVE IS OPEN}" with the linear trend removed.}
    \label{fig:echoclip_promptdiff}
\end{figure}
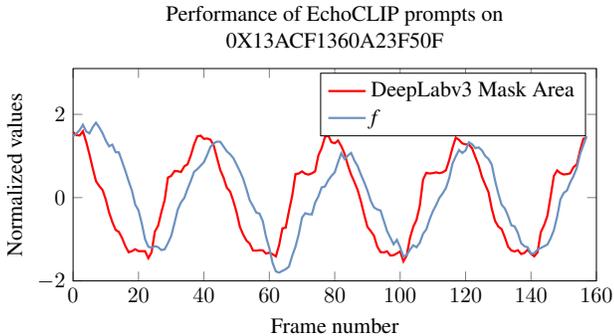

We tried different combinations of prompts. Differentiating the result of the prompt ``\textit{THE MITRAL VALVE IS CLOSED}" from the prompt ``\textit{THE MITRAL VALVE IS OPEN}" and then integrating this signal yields a convincing plot that is consistent with that of other models and with human labels on many examples, as shown on Figure~\ref{fig:echoclip_promptdiff}.

\vspace{-5pt}
Due to the novelty of this approach and the lack of previous scores to compare against, we report the results in the supplementary material. Note that we find the same trend when comparing to human labels.

\subsection{Results}
\label{sub:results}
\vspace{-10pt}
\begin{table}[h!]
\centering
\resizebox{0.85\linewidth}{!}{
\begin{tabular}{lcccc}
\toprule
&\#& \aFDED & \aFDES \\
\midrule
DeepLabv3$^*$\cite{ouyang2020echonet} & 40M & 8.63 & 3.66 \\
UVT$_R$\cite{Reynaud2021aFD} & 347M &7.88 & 2.86\\
UVT$_M$\cite{Reynaud2021aFD} & 347M & 7.17 & 3.35 \\
EchoGNN\cite{mokhtari2022echognn} & 1.7M& 3.68 & 4.15 \\

DRNNEcho\cite{dezaki2017deep} & (\(>\)10M)\tablefootnote{Authors claimed to use ResNets with 126 layers and some LSTM on top, so this number of parameters is very approximated.} & 3.7 & 4.1 \\
\midrule

\ourmodel & 0.72M & \textbf{2.72} & \textbf{2.83} \\

\ourmodel$^*$ & 0.72M & \underline{2.77} & \underline{2.83}\\
\bottomrule
\end{tabular}
}
\caption{Number of parameters and aFD obtained using various segmentators. All results reported in the table are computed on a subset\tablefootnote{The authors removed some complicated samples to compute their aFD prediction. } of the EchoNet-Dynamic dataset. We also provide results of the entire test set ($^*$).
\label{tab:afd}
	\textbf{Best results - in bold}, \underline{second best - underlined}.
\vspace{-10pt}}
\end{table}

\paragraph{Comparison to real-data trained methods} 
Table~\ref{tab:afd} shows that \ourmodel~outperforms almost all compared methods in setups where the real data is accessible to train on directly.
However, the scores for each experiment are close. Additionally, it can be observed that training on synthetic data requires 3 to 4 times fewer weights. This suggests that the intrinsic complexity of the synthetic dataset concerning the tasks we are performing is lower, in similar proportions. However, this hypothesis should be taken cautiously, and quantifying each dataset's dimensionality would require more rigorous exploration. Without delving into such deep considerations, it can be noted that when the model trained on synthetic data is evaluated on real data, most of the total error is concentrated on a small portion of the test examples (see supplementary materials for details).
\vspace{-5pt}
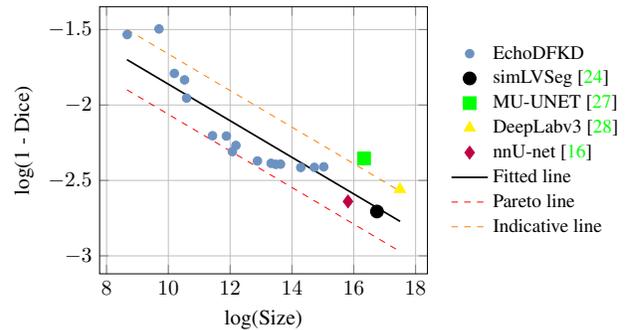
\begin{figure}[h]
    \centering
    \resizebox{\linewidth}{!}{    \begin{tikzpicture}
    \begin{axis}[
        width=7cm, height=6cm,
        xlabel={log(Size)},
        ylabel={log(1 - Dice)},
        grid=major,
        legend style={
            at={(1.05,0.5)},
            anchor=west,
            draw=none,
            font=\small,
        },
        legend cell align={left}
    ]

    \addplot[
        only marks,
        mark=*,
        color=bluefig
    ] table {
        x    y
8.6692273 -1.5333877
9.6979997 -1.4958755
10.1942523 -1.7908473
10.5244134 -1.8336489
10.5908429 -1.9544327
11.4284236 -2.2037633
11.8777421 -2.2058268
12.1866502 -2.2680334
12.0728669 -2.3089105
12.8827270 -2.3710695
13.3242538 -2.3866626
13.6294905 -2.3922138
13.4803189 -2.3934659
14.2842139 -2.4143354
14.7240275 -2.4134842
15.0284513 -2.4097077
    };
    \addlegendentry{EchoDFKD}

    \addplot[
        only marks,
        mark=*, 
        color=black,
        mark size=3pt
    ] coordinates {(16.7509619, -2.7060522)};
    \addlegendentry{simLVSeg~\cite{maani2024simlvseg}}

    \addplot[
        only marks,
        mark=square*,
        color=green,
        mark size=3pt
    ] coordinates {(16.3315928, -2.3538783)};
    \addlegendentry{MU-UNET~\cite{muldoon2023lightweight}}

    \addplot[
        only marks,
        mark=triangle*,
        color=yellow,
        mark size=3pt
    ] coordinates {(17.4953492, -2.5587685)};
    \addlegendentry{DeepLabv3~\cite{ouyang2020echonet}}

    \addplot[
        only marks,
        mark=diamond*,
        color=purple,
        mark size=3pt
    ] coordinates {(15.81292826, -2.63945741)};
    \addlegendentry{nnU-net~\cite{isensee2021nnu}}

    \addplot[
        thick,
        color=black,
        domain=8.66:17.5,        
        samples=100
    ] {(-0.12126752184422228 * x - 0.649230393007148)};
    \addlegendentry{Fitted line}

    \addplot[
        dashed,
        color=red,
        domain=8.66:17.5,        
        samples=100
    ] {(-0.12126752184422228 * x - 0.649230393007148 - 0.2)};
    \addlegendentry{Pareto line}

    \addplot[
        dashed,
        color=orange,
        domain=8.66:17.5,        
        samples=100
    ] {(-0.12126752184422228 * x - 0.649230393007148 + 0.2)};
    \addlegendentry{Indicative line}

    \end{axis}
    \end{tikzpicture}} \vspace{-20pt}
    \caption{Log-log display of the performances of different \ourmodel~ models as a function of model size, as well as several models from the literature. The performances of the EchoDFKD models reach a ceiling as they approach their teacher's performance.}\vspace{-5pt}
    \label{fig:loglog}
\end{figure}

The deep learning community has established that the evolution of model performance scores as a function of their size is generally well-described by exponential scaling laws, at least in a regime preceding the saturation imposed by the dataset size or other factors \cite{rosenfeld2019constructive,sharma2022scaling}. We therefore examined, on the one hand, $-\log($\aFDED + \aFDES$)$ as a function of $\log(\text{Model\_size})$, and, on the other hand, $\log(1-\meanIoU)$ as a function of $\log(\text{Model\_size})$.  These analyses were conducted both for evaluation against human annotations and using EchoCLIP rewards (see plots in the supplementary material). We observed clear linear trends up to the largest models, where performance begins to stagnate as referred in Figure~\ref{fig:loglog}.

\noindent\textbf{FLOPS}~~In evaluating our model's efficiency, we consider the number of FLOPS required for inference. EchoNet-Dynamic's DeepLabv3 operates at 7.84 GFLOPS, reflecting its computational intensity and resource demands. In contrast, our models demonstrate significantly lower computational requirements, with GFLOPS values of 0.25 and 1.56, respectively, for the results reported in Table~\ref{tab:afd}.

\section{Limitations and future work}
\vspace{-5pt}
Our experiments show that it is possible to obtain results close to the SOTA in the segmentation stage and to outperform the SOTA in the frame identification stage with tiny models. Training models on synthetic data rather than real data also yields competitive results. The fact that we managed to surpass the SOTA with so few parameters may be due to our model producing smoothed outputs precisely because of its compression and its recurrent nature. Applying low-pass filters to a signal can sometimes improve peak detection, and we have probably reproduced a similar effect.

It is worth noting that we use ConvLSTMs in a mode where there is only one pass, so they have no hindsight and must make decisions in real-time, frame by frame, as soon as they arrive. This mainly implies poor performance in the very first frames (see supplementary materials), as the model likely needs to grasp the situation. It also means that it is more difficult for the model to identify the ES and ED frames since it cannot know if the next frame will show a stagnation or a change in the observed motion.

In this work, we only used 10,000 artificial examples of EchoNet-Synthetic, the generative model being recent and the generation of new examples time-consuming. In addition, we only used one teacher model rather than aggregating outputs from various models. It would be interesting to extend the data-free experiment into a large-scale one, using a cohort of teachers that could include models trained on other datasets and distillation dataset much more significant than 10,000 examples. 

Our results suggest that it does not take much complexity to model the variables of interest from standard ultrasound scans.
Moreover, the trend observed in model performance when varying the number of weights suggests that models may quickly reach a glass ceiling imposed by a certain vagueness around input standardization conventions and the limits of human labels.
Last but not least, we must not lose sight of the fact that the tasks used as benchmarks for the many papers published on EchoNet-Dynamic are part of an approach that ultimately consists of evaluating LVEF. Ultrasound scans likely contain other interesting information from a medical point of view, which would be of more interest than scraping a few decimal points on a conventional score, and this is particularly true of examples where the video clearly shows patients that have other significant dysfunctions, as it was pointed out by~\cite{ouyang2020echonet}.

This has several critical implications. Firstly, it implies that, given the simplicity of the inputs and targets, choosing one architecture over another is unlikely to impact the results obtained significantly. Multiplying work by trying out a whole panoply of different architectures, which are rarely original but adapted from models already well established in computer vision, is therefore likely to have limited scientific interest in this dataset and these tasks in the future.

If researchers want to continue improving model scores on these tasks, seeking richer or more precise targets will likely be necessary. In this paper, we proposed an approach involving a large model as an evaluator. The large model does not have its own style, as it has seen data processed by various practitioners, and the fact that it has seen data in multiple modalities simultaneously may give it the ability to access, from ultrasound scans, information that is difficult for humans to perceive with this single modality. We generally seek information from other modalities; what happens along the third dimension is usually analyzed with a complementary 2-chamber view. However, the EchoCLIP verdicts remain pretty noisy, implying that many examples would be needed to discriminate between models reliably. EchoCLIP-like models may improve, and the approaches we explored with synthetic data for the training phase could be extended to the evaluation phase. A more straightforward and traditional way to obtain rich targets would be to collect multiple human labels for each evaluation example and combine these labels. It should also be noted that less downsampled data would also provide more precise labels. We have seen this with the choice of ES and ED frames, which makes less sense for videos with a low sampling rate. Still, more generally, downsampling videos to 112$\times$112 pixels compresses much information on the ability to capture contours.


\section{Conclusion}
\vspace{-5pt}
In this work, we have demonstrated the potential of using DFKD in video segmentation applied to cardiac ultrasound. First, we highlighted the challenges of using real data to train our model and how, with synthetic data, KD can overcome data access issues and effectively train a much lighter network. Second, we proposed a robust method to compute the aFD scores accurately. This method only requires an estimate of the inner left ventricle mask and thus is very easily transferrable. 
Finally, we have demonstrated a method for evaluating a model's ability to segment medical images using a large model, which gives results consistent with an evaluation of the fully labeled test set. Unlike the latter approach, ours does not require human labels and challenges the need to produce segmentation labels on test sets. A possibility to generalize this approach to training is using reinforcement learning to take advantage of feedback on mask quality.

Our work has shown possibilities in knowledge dissemination that circumvent limits imposed by the need to keep data private, as well as the possibility of extracting knowledge from heavy models on the best-understood part of the latent data space. We have also shown that unconventional evaluation methods can avoid costly and time-consuming human segmentation labels by taking advantage of a large existing model.  We hope that by developing these methods, it will be possible to better share the significant but scattered efforts made in medical imaging.

{\small
\bibliographystyle{ieee_fullname}
\bibliography{egbib}

\begin{thebibliography}{10}\itemsep=-1pt

\bibitem{aytar2016soundnet}
Yusuf Aytar, Carl Vondrick, and Antonio Torralba.
\newblock Soundnet: Learning sound representations from unlabeled video.
\newblock {\em Advances in neural information processing systems}, 29, 2016.

\bibitem{azizi2023synthetic}
Shekoofeh Azizi, Simon Kornblith, Chitwan Saharia, Mohammad Norouzi, and David~J Fleet.
\newblock Synthetic data from diffusion models improves imagenet classification.
\newblock {\em arXiv preprint arXiv:2304.08466}, 2023.

\bibitem{batool2023ejection}
Samana Batool, Imtiaz~Ahmad Taj, and Mubeen Ghafoor.
\newblock Ejection fraction estimation from echocardiograms using optimal left ventricle feature extraction based on clinical methods.
\newblock {\em Diagnostics}, 13(13):2155, 2023.

\bibitem{bommasani2021opportunities}
Rishi Bommasani, Drew~A Hudson, Ehsan Adeli, Russ Altman, Simran Arora, Sydney von Arx, Michael~S Bernstein, Jeannette Bohg, Antoine Bosselut, Emma Brunskill, et~al.
\newblock On the opportunities and risks of foundation models.
\newblock {\em arXiv preprint arXiv:2108.07258}, 2021.

\bibitem{Camporese2020action}
Guglielmo Camporese, Pasquale Coscia, Antonino Furnari, Giovanni~Maria Farinella, and Lamberto Ballan.
\newblock Knowledge distillation for action anticipation via label smoothing.
\newblock {\em arXiv preprint arXiv:2004.07711}, 2020.

\bibitem{caruana1997multitask}
Rich Caruana.
\newblock Multitask learning.
\newblock {\em Machine learning}, 28:41--75, 1997.

\bibitem{caruana2004ensemble}
Rich Caruana, Alexandru Niculescu-Mizil, Geoff Crew, and Alex Ksikes.
\newblock Ensemble selection from libraries of models.
\newblock In {\em Proceedings of the twenty-first international conference on Machine learning}, page~18, 2004.

\bibitem{chen2019data}
Hanting Chen, Yunhe Wang, Chang Xu, Zhaohui Yang, Chuanjian Liu, Boxin Shi, Chunjing Xu, Chao Xu, and Qi Tian.
\newblock Data-free learning of student networks.
\newblock In {\em Proceedings of the IEEE/CVF international conference on computer vision}, pages 3514--3522, 2019.

\bibitem{christensen2024echoclip}
Matthew Christensen, Milos Vukadinovic, Neal Yuan, and David Ouyang.
\newblock Vision--language foundation model for echocardiogram interpretation.
\newblock {\em Nature Medicine}, pages 1--8, 2024.

\bibitem{dezaki2017deep}
Fatemeh~Taheri Dezaki, Neeraj Dhungel, Amir~H Abdi, Christina Luong, Teresa Tsang, John Jue, Ken Gin, Dale Hawley, Robert Rohling, and Purang Abolmaesumi.
\newblock Deep residual recurrent neural networks for characterisation of cardiac cycle phase from echocardiograms.
\newblock In {\em Deep Learning in Medical Image Analysis and Multimodal Learning for Clinical Decision Support: Third International Workshop, DLMIA 2017, and 7th International Workshop, ML-CDS 2017, Held in Conjunction with MICCAI 2017, Qu{\'e}bec City, QC, Canada, September 14, Proceedings 3}, pages 100--108. Springer, 2017.

\bibitem{fan2020learning}
Junsong Fan, Zhaoxiang Zhang, Chunfeng Song, and Tieniu Tan.
\newblock Learning integral objects with intra-class discriminator for weakly-supervised semantic segmentation.
\newblock In {\em Proceedings of the IEEE/CVF conference on computer vision and pattern recognition}, pages 4283--4292, 2020.

\bibitem{fazry2022hierarchical}
Lhuqita Fazry, Asep Haryono, Nuzulul~Khairu Nissa, Naufal~Muhammad Hirzi, Muhammad~Febrian Rachmadi, Wisnu Jatmiko, et~al.
\newblock Hierarchical vision transformers for cardiac ejection fraction estimation.
\newblock In {\em 2022 7th International Workshop on Big Data and Information Security (IWBIS)}, pages 39--44. IEEE, 2022.

\bibitem{guan2018said}
Melody Guan, Varun Gulshan, Andrew Dai, and Geoffrey Hinton.
\newblock Who said what: Modeling individual labelers improves classification.
\newblock In {\em Proceedings of the AAAI conference on artificial intelligence}, volume~32, 2018.

\bibitem{he2023synthetic}
Ruifei He, Shuyang Sun, Xin Yu, Chuhui Xue, Wenqing Zhang, Philip Torr, Song Bai, and XIAOJUAN QI.
\newblock Is synthetic data from generative models ready for image recognition?
\newblock In {\em The Eleventh International Conference on Learning Representations}, 2023.

\bibitem{hinton2015distilling}
Geoffrey Hinton, Oriol Vinyals, and Jeff Dean.
\newblock Distilling the knowledge in a neural network.
\newblock {\em arXiv preprint arXiv:1503.02531}, 2015.

\bibitem{isensee2021nnu}
Fabian Isensee, Paul~F Jaeger, Simon~AA Kohl, Jens Petersen, and Klaus~H Maier-Hein.
\newblock nnu-net: a self-configuring method for deep learning-based biomedical image segmentation.
\newblock {\em Nature methods}, 18(2):203--211, 2021.

\bibitem{kazemi2020deep}
Mohammad~Mahdi Kazemi~Esfeh, Christina Luong, Delaram Behnami, Teresa Tsang, and Purang Abolmaesumi.
\newblock A deep bayesian video analysis framework: towards a more robust estimation of ejection fraction.
\newblock In {\em International Conference on Medical Image Computing and Computer-Assisted Intervention}, pages 582--590. Springer, 2020.

\bibitem{lan2023vision}
Shiyi Lan, Xitong Yang, Zhiding Yu, Zuxuan Wu, Jose~M Alvarez, and Anima Anandkumar.
\newblock Vision transformers are good mask auto-labelers.
\newblock In {\em Proceedings of the IEEE/CVF Conference on Computer Vision and Pattern Recognition}, pages 23745--23755, 2023.

\bibitem{leclerc2019deep}
Sarah Leclerc, Erik Smistad, Joao Pedrosa, Andreas {\O}stvik, Frederic Cervenansky, Florian Espinosa, Torvald Espeland, Erik Andreas~Rye Berg, Pierre-Marc Jodoin, Thomas Grenier, et~al.
\newblock Deep learning for segmentation using an open large-scale dataset in 2d echocardiography.
\newblock {\em IEEE transactions on medical imaging}, 38(9):2198--2210, 2019.

\bibitem{li2019recurrent}
Ming Li, Weiwei Zhang, Guang Yang, Chengjia Wang, Heye Zhang, Huafeng Liu, Wei Zheng, and Shuo Li.
\newblock Recurrent aggregation learning for multi-view echocardiographic sequences segmentation.
\newblock In {\em Medical Image Computing and Computer Assisted Intervention--MICCAI 2019: 22nd International Conference, Shenzhen, China, October 13--17, 2019, Proceedings, Part II 22}, pages 678--686. Springer, 2019.

\bibitem{li2020knowledge}
Wei-Hong Li and Hakan Bilen.
\newblock Knowledge distillation for multi-task learning.
\newblock {\em arXiv preprint arXiv:2007.06889}, 2020.

\bibitem{lopes2017data}
Raphael~Gontijo Lopes, Stefano Fenu, and Thad Starner.
\newblock Data-free knowledge distillation for deep neural networks.
\newblock {\em arXiv preprint arXiv:1710.07535}, 2017.

\bibitem{ma2024segment}
Jun Ma, Yuting He, Feifei Li, Lin Han, Chenyu You, and Bo Wang.
\newblock Segment anything in medical images.
\newblock {\em Nature Communications}, 15(1):654, 2024.

\bibitem{maani2024simlvseg}
Fadillah Maani, Asim Ukaye, Nada Saadi, Numan Saeed, and Mohammad Yaqub.
\newblock Simlvseg: Simplifying left ventricular segmentation in 2d+time echocardiograms with self- and weakly-supervised learning, 2024.

\bibitem{mokhtari2022echognn}
Masoud Mokhtari, Teresa Tsang, Purang Abolmaesumi, and Renjie Liao.
\newblock Echognn: explainable ejection fraction estimation with graph neural networks.
\newblock In {\em International Conference on Medical Image Computing and Computer-Assisted Intervention}, pages 360--369. Springer, 2022.

\bibitem{muhtaseb2022echocotr}
Rand Muhtaseb and Mohammad Yaqub.
\newblock Echocotr: Estimation of the left ventricular ejection fraction from spatiotemporal echocardiography.
\newblock In {\em International Conference on Medical Image Computing and Computer-Assisted Intervention}, pages 370--379. Springer, 2022.

\bibitem{muldoon2023lightweight}
Meghan Muldoon and Naimul Khan.
\newblock Lightweight and interpretable left ventricular ejection fraction estimation using mobile u-net.
\newblock In {\em 2023 IEEE 20th International Symposium on Biomedical Imaging (ISBI)}, pages 1--5. IEEE, 2023.

\bibitem{ouyang2020echonet}
David Ouyang, Bryan He, Amirata Ghorbani, Neal Yuan, Joseph Ebinger, Curtis~P Langlotz, Paul~A Heidenreich, Robert~A Harrington, David~H Liang, Euan~A Ashley, et~al.
\newblock Video-based ai for beat-to-beat assessment of cardiac function.
\newblock {\em Nature}, 580(7802):252--256, 2020.

\bibitem{papernot2016semi}
Nicolas Papernot, Mart{\'\i}n Abadi, Ulfar Erlingsson, Ian Goodfellow, and Kunal Talwar.
\newblock Semi-supervised knowledge transfer for deep learning from private training data.
\newblock {\em arXiv preprint arXiv:1610.05755}, 2016.

\bibitem{papernot2016distillation}
Nicolas Papernot, Patrick McDaniel, Xi Wu, Somesh Jha, and Ananthram Swami.
\newblock Distillation as a defense to adversarial perturbations against deep neural networks.
\newblock In {\em 2016 IEEE Symposium on Security and Privacy (SP)}, pages 582--597, Los Alamitos, CA, USA, may 2016. IEEE Computer Society.

\bibitem{radford2021clip}
Alec Radford, Jong~Wook Kim, Chris Hallacy, Aditya Ramesh, Gabriel Goh, Sandhini Agarwal, Girish Sastry, Amanda Askell, Pamela Mishkin, Jack Clark, et~al.
\newblock Learning transferable visual models from natural language supervision.
\newblock In {\em International conference on machine learning}, pages 8748--8763. PMLR, 2021.

\bibitem{rajpurkar2022ai}
Pranav Rajpurkar, Emma Chen, Oishi Banerjee, and Eric~J Topol.
\newblock Ai in health and medicine.
\newblock {\em Nature medicine}, 28(1):31--38, 2022.

\bibitem{reynaud2024EchoNetsynthetic}
Hadrien Reynaud, Qingjie Meng, Mischa Dombrowski, Arijit Ghosh, Thomas Day, Alberto Gomez, Paul Leeson, and Bernhard Kainz.
\newblock Echonet-synthetic: Privacy-preserving video generation for safe medical data sharing.
\newblock {\em arXiv preprint arXiv:2406.00808}, 2024.

\bibitem{reynaud2023feature}
Hadrien Reynaud, Mengyun Qiao, Mischa Dombrowski, Thomas Day, Reza Razavi, Alberto Gomez, Paul Leeson, and Bernhard Kainz.
\newblock Feature-conditioned cascaded video diffusion models for precise echocardiogram synthesis.
\newblock In {\em International Conference on Medical Image Computing and Computer-Assisted Intervention}, pages 142--152. Springer, 2023.

\bibitem{Reynaud2021aFD}
Hadrien Reynaud, Athanasios Vlontzos, Benjamin Hou, Arian Beqiri, Paul Leeson, and Bernhard Kainz.
\newblock Ultrasound video transformers for cardiac ejection fraction estimation.
\newblock In {\em MICCAI}. Springer, 2021.

\bibitem{ronneberger2015u}
Olaf Ronneberger, Philipp Fischer, and Thomas Brox.
\newblock U-net: Convolutional networks for biomedical image segmentation.
\newblock In {\em Medical image computing and computer-assisted intervention--MICCAI 2015: 18th international conference, Munich, Germany, October 5-9, 2015, proceedings, part III 18}, pages 234--241. Springer, 2015.

\bibitem{rosenfeld2019constructive}
{Jonathan S.} Rosenfeld, Amir Rosenfeld, Yonatan Belinkov, and Nir Shavit.
\newblock A constructive prediction of the generalization error across scales.
\newblock In {\em International Conference on Learning Representations (ICLR) 2020}, 2020.

\bibitem{saeed2022contrastive}
Mohamed Saeed, Rand Muhtaseb, and Mohammad Yaqub.
\newblock Is contrastive learning suitable for left ventricular segmentation in echocardiographic images?
\newblock {\em arXiv}, 2022.

\bibitem{shamshad2023transformers}
Fahad Shamshad, Salman Khan, Syed~Waqas Zamir, Muhammad~Haris Khan, Munawar Hayat, Fahad~Shahbaz Khan, and Huazhu Fu.
\newblock Transformers in medical imaging: A survey.
\newblock {\em Medical Image Analysis}, 88:102802, 2023.

\bibitem{sharma2022scaling}
Utkarsh Sharma and Jared Kaplan.
\newblock Scaling laws from the data manifold dimension.
\newblock {\em Journal of Machine Learning Research}, 23(9):1--34, 2022.

\bibitem{shi2015convolutional}
Xingjian Shi, Zhourong Chen, Hao Wang, Dit-Yan Yeung, Wai-Kin Wong, and Wang-chun Woo.
\newblock Convolutional lstm network: A machine learning approach for precipitation nowcasting.
\newblock {\em Advances in neural information processing systems}, 28, 2015.

\bibitem{shi2017deep}
Xingjian Shi, Zhihan Gao, Leonard Lausen, Hao Wang, Dit-Yan Yeung, Wai-kin Wong, and Wang-chun Woo.
\newblock Deep learning for precipitation nowcasting: A benchmark and a new model.
\newblock {\em Advances in neural information processing systems}, 30, 2017.

\bibitem{szatkowski2024adapt}
Filip Szatkowski, Mateusz Pyla, Marcin Przewiezlikowski, Sebastian Cygert, Bartlomiej Twardowski, and Tomasz Trzcinski.
\newblock Adapt your teacher: Improving knowledge distillation for exemplar-free continual learning.
\newblock In {\em Proceedings of the IEEE/CVF Winter Conference on Applications of Computer Vision}, pages 1977--1987, 2024.

\bibitem{tan2024large}
Zhen Tan, Alimohammad Beigi, Song Wang, Ruocheng Guo, Amrita Bhattacharjee, Bohan Jiang, Mansooreh Karami, Jundong Li, Lu Cheng, and Huan Liu.
\newblock Large language models for data annotation: A survey.
\newblock {\em arXiv preprint arXiv:2402.13446}, 2024.

\bibitem{warfield2004simultaneous}
Simon~K Warfield, Kelly~H Zou, and William~M Wells.
\newblock Simultaneous truth and performance level estimation (staple): an algorithm for the validation of image segmentation.
\newblock {\em IEEE transactions on medical imaging}, 23(7):903--921, 2004.

\bibitem{yao2021filip}
Lewei Yao, Runhui Huang, Lu Hou, Guansong Lu, Minzhe Niu, Hang Xu, Xiaodan Liang, Zhenguo Li, Xin Jiang, and Chunjing Xu.
\newblock Filip: Fine-grained interactive language-image pre-training.
\newblock In {\em International Conference on Learning Representations}, 2021.

\bibitem{yi2023simple}
Muyang Yi, Quan Cui, Hao Wu, Cheng Yang, Osamu Yoshie, and Hongtao Lu.
\newblock A simple framework for text-supervised semantic segmentation.
\newblock In {\em Proceedings of the IEEE/CVF Conference on Computer Vision and Pattern Recognition}, pages 7071--7080, 2023.

\bibitem{yoo2019knownledgeextraction}
Jaemin Yoo, Minyong Cho, Taebum Kim, and U Kang.
\newblock Knowledge extraction with no observable data.
\newblock In H. Wallach, H. Larochelle, A. Beygelzimer, F. d\textquotesingle Alch\'{e}-Buc, E. Fox, and R. Garnett, editors, {\em Advances in Neural Information Processing Systems}, volume~32. Curran Associates, Inc., 2019.

\bibitem{you2017learning}
Shuyang You, Chang Xu, Chao Xu, and Dacheng Tao.
\newblock Learning from multiple teacher networks.
\newblock In {\em Proceedings of the 23rd ACM SIGKDD International Conference on Knowledge Discovery and Data Mining}, pages 1285--1294, 2017.

\bibitem{zhang2018fully}
Jeffrey Zhang, Sravani Gajjala, Pulkit Agrawal, Geoffrey~H Tison, Laura~A Hallock, Lauren Beussink-Nelson, Mats~H Lassen, Eugene Fan, Mandar~A Aras, ChaRandle Jordan, et~al.
\newblock Fully automated echocardiogram interpretation in clinical practice: feasibility and diagnostic accuracy.
\newblock {\em Circulation}, 138(16):1623--1635, 2018.

\end{thebibliography}
}

\newpage~\newpage

\twocolumn[
\begin{center}
    \Large\textbf{Supplementary material of ``EchoDFKD: Data-Free Knowledge Distillation for Cardiac Ultrasound Segmentation using Synthetic Data"}\\~\\~
\end{center}
]
\setcounter{section}{0}
\section*{Introduction}
In this supplementary material, we provide:

\begin{itemize}
\setlength\itemsep{0.1em}
 \item \ref{sec:currupted}: Details regarding the varied sampling rates and different image qualities present in the dataset, including examples of corrupted clips.
 \item \ref{sec:proof}: Proof that the model outputs can be extremely close to the mean annotation of the annotator while being less noisy.
 \item \ref{sec:ext_results}: Extensive results due to the novelty of our approach and the lack of previous scores to compare against. The raw results can be found under \texttt{echoclip.csv}
 \item \ref{sec:small_portion}: When the model trained on synthetic data is evaluated on real data, most of the total error is concentrated on a small portion of the test examples.
 \item \ref{sec:scaling_law}: Plots of $\log(scores)$ as a function of $\log(\text{Model\_size})$ for evaluation against human annotations and via EchoCLIP rewards. 
 \item \ref{sec:first_frames}: An illustration of the poor performance in the very first frames, as noted in the main paper.
 \item \ref{sec:multi}: An extension of \ourmodel~to a multi-teacher setting where the right ventricle segmentation is also learned.
 \item \ref{sec:camus}: \ourmodel~inference on CAMUS dataset
\end{itemize}

\section{Corrupted examples}
\label{sec:currupted}
\begin{figure}[h]
    \centering
    \includegraphics[width=\linewidth]{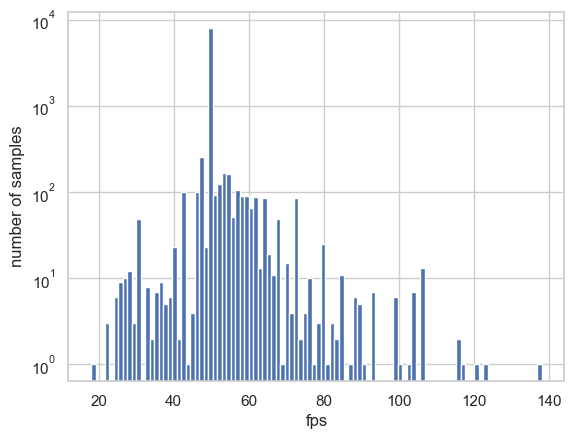}
    \caption{Distribution of sampling rates in the EchoNet-Dynamic~\cite{ouyang2020echonet} dataset.}
    \label{fig:sr_distribution}
\end{figure}

As mentioned in Subsection 3.1. of the main paper, the EchoNet-Dynamic~\cite{ouyang2020echonet} dataset is very heterogeneous in terms of sampling rate (Figure~\ref{fig:sr_distribution}), or image quality (Figure~\ref{fig:foreshortening}).

\begin{figure}
 \centering
 0X39348579B2E55470\\\includegraphics[width=0.75\linewidth]{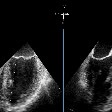}

 0X3693781992586497\\\includegraphics[width=0.75\linewidth]{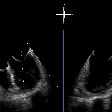}
 
 0X790C871B162806D2\\\includegraphics[width=0.75\linewidth]{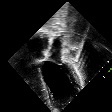}
 \caption{3 most corrupted examples.}
 \label{fig:corrupted}
\end{figure}

The most corrupted examples are 0X39348579B2E55470, 0X3693781992586497, and 0X790C871B162806D2, as displayed in Figure~\ref{fig:corrupted}.

Additionally, we include in \texttt{corrupted.csv} different lists of corrupted examples for the following reasons:
\begin{itemize}
\setlength\itemsep{0.1em}
\item Videos that are manifestly corrupted (see Figure~\ref{fig:corrupted}).
\item Labeling masks that are corrupted due to issues with \cite{ouyang2020echonet}'s function fail to load labels properly for multi-labeled examples.
\item Cases where the End-Systolic (ES) and End-Diastolic (ED) frames are too close together (differences close to one or even one in some examples).
\end{itemize}

\section{Derivation of theoretical bounds for model scores in the function of intra-annotator scores}
\label{sec:proof}

It seems that two slightly different definitions of intra-annotator standard deviation currently coexist in the literature. One considers the deviation between the values of a second annotation session and the first session, and the other considers the deviation between the values from one of the two sessions and a merged value from the two sessions (typically the mean). Here, since the RMSE reported in CAMUS is rather high compared to what we can observe from some models, we can infer that they used the first convention.

Consider two rounds of annotations, \(Z_1\) and \(Z_2\). We assume:
\[
Z_1 = X_1 + Y
\]
\[
Z_2 = X_2 + Y
\]
with \(X_1\) and \(X_2\) centered, i.i.d. (which is not very realistic but simplifies the derivations a lot), and \(Y\) being the latent truth (or, at least, the tendential value we would find with a lot of rounds).

The RMSE of the second round as an estimator of the first is :

\[
\begin{aligned}
\text{RMSE}(Z_2, Z_1) &= \sqrt{\mathbb{E}[(Z_2 - Z_1)^2]} \\
                       &= \sqrt{\mathbb{E}[(X_2 - X_1)^2]} \\
                       &= \sqrt{2\sigma_X^2} (\text{since \(X_1\) and \(X_2\) are independent}) \\
                       &= \sqrt{2} \cdot \sigma_X
\end{aligned}
\]

Now, the RMSE of a perfect model that would output Y, compared with a target obtained with a single annotation per example, is

\[
\begin{aligned}
\text{RMSE}(Y, Z) &= \sqrt{\mathbb{E}[(Z - Y)^2]} \\
                     &= \sqrt{\mathbb{E}[(X + Y - Y)^2]} \\
                     &= \sqrt{\mathbb{E}[X^2]} \\
                     &= \sigma_X
\end{aligned}
\]

We get:

\[
\begin{aligned}
\text{RMSE}(Z_2, Z_1) &= \sqrt{2} \cdot \text{RMSE}(Z_2, Y)
\end{aligned}
\]

Thus, the RMSE of \( Z_2 \) as an estimate of \( Z_1 \) is \( \sqrt{2} \) times the RMSE of \( Z_2 \) with respect to \( Y \).

CAMUS reports an intra-annotator std of 5.7. The theoretical lower bound of model performance is thus 4.03

On EchoNet-Dynamic, EchoCoTr reports an RMSE of 5.17

We can also look at the theoretical bound for correlation.

The correlation between \(Z\) and \(Y\) is :

\begin{align*}
\rho_{Z, Y} &= \frac{\text{Cov}(Z, Y)}{\sigma_{Z} \sigma_Y} \\
&= \frac{\text{Cov}(Y + X, Y)}{\sigma_{Z} \sigma_Y} \\
&= \frac{\sigma_Y^2}{\sigma_{Z} \sigma_Y} \\
&= \frac{\sigma_Y}{\sqrt{\sigma_Y^2 + \sigma_{X}^2}}
\end{align*}

Next, the intra-annotator correlation, i.e. the correlation between \(Z_1\) and \(Z_2\) is :

\begin{align*}
\rho_{Z_1, Z_2} &= \frac{\text{Cov}(Z_1, Z_2)}{\sigma_{Z_1} \sigma_{Z_2}} \\
&= \frac{\text{Cov}(Y + X_1, Y + X_2)}{\sigma_{Z_1} \sigma_{Z_2}} \\
&= \frac{\sigma_Y^2}{\sigma_{Z_1} \sigma_{Z_2}} \\
&= \frac{\sigma_Y^2}{\sigma_{Z_1}^2} \\
&= \frac{\sigma_Y^2}{\sigma_Y^2 + \sigma_{X_1}^2} \\
\end{align*}

Finally, we have :
\[
\rho_{Z_1, Y} = \sqrt{\rho_{Z_1, Z_2}}
\]

The best correlation coefficient that can be achieved between the model and the labeler, if only one annotation per example is available, is therefore \( \sqrt{\rho_{Z_1, Z_2}} \).

CAMUS reports an intra-annotator correlation of 0.801. Thus, the theoretical upper bound is 0.895.
\cite{batool2023ejection}, for instance, report a correlation of 0.78.

EchoCoTr doesn't provide a correlation. However, they report their R$^2$, which should be lower. After a linear regression between outputs and targets, the quadratic errors sum will be smaller (one would add two parameters that are allowed to fit the data used for evaluation); thus,  R$^2$ of new outputs will be higher, and it will be equal to the correlation coefficient.
Their squared correlation coefficient is higher than the reported R$^2$, making them close to the theoretical bound.

We also took an interest in the theoretical limit of a model's aFD score. This corresponds to the MAE of the selected frame number, and, through reasoning similar to the previous ones, it represents the average error of an annotator compared to a reference frame. We do not have access to this reference frame. Instead, we added an additional round of annotation by labeling over a thousand examples ourselves. This gives us an empirical distribution of \( Z_1 - Z_2 \), where \( Z_1 \) and \( Z_2 \) are the two rounds of labeling. By setting \( Z_1 = Y + X_1 \) and \( Z_2 = Y + X_2 \), with \( Y \) as the reference frame, and \( X_1 \) and \( X_2 \) iid, this reduces to \( X_1 - X_2 \), which distribution is obtained from that of \( X \) by convolution. Another practical assumption is to suppose that the distribution of X is the sum of a uniform distribution, which represents an annotator's abandonment when faced with a particularly degraded example (which happens very rarely, maybe once in several hundred examples, and results in discrepancies between two annotators that can reach up to fifty frames), and a symmetric distribution over a smaller support, which represents variations due to an annotator's lack of precision or a sequence of indistinct frames (resulting in discrepancies that rarely exceed nine or ten frames). We find that a Laplace distribution provides excellent log-likelihood after convolution by fitting different types of discrete distributions for the small support term. The expected distribution value obtained for \( \|X\|_1 \) is approximately 2.0 for the ES frame, and 2.4 for the ED frame.

\section{Extensive results}
\label{sec:ext_results}
\begin{figure}
    \centering
    High ``foreshortening" response\\
    \includegraphics[width=0.3\linewidth]{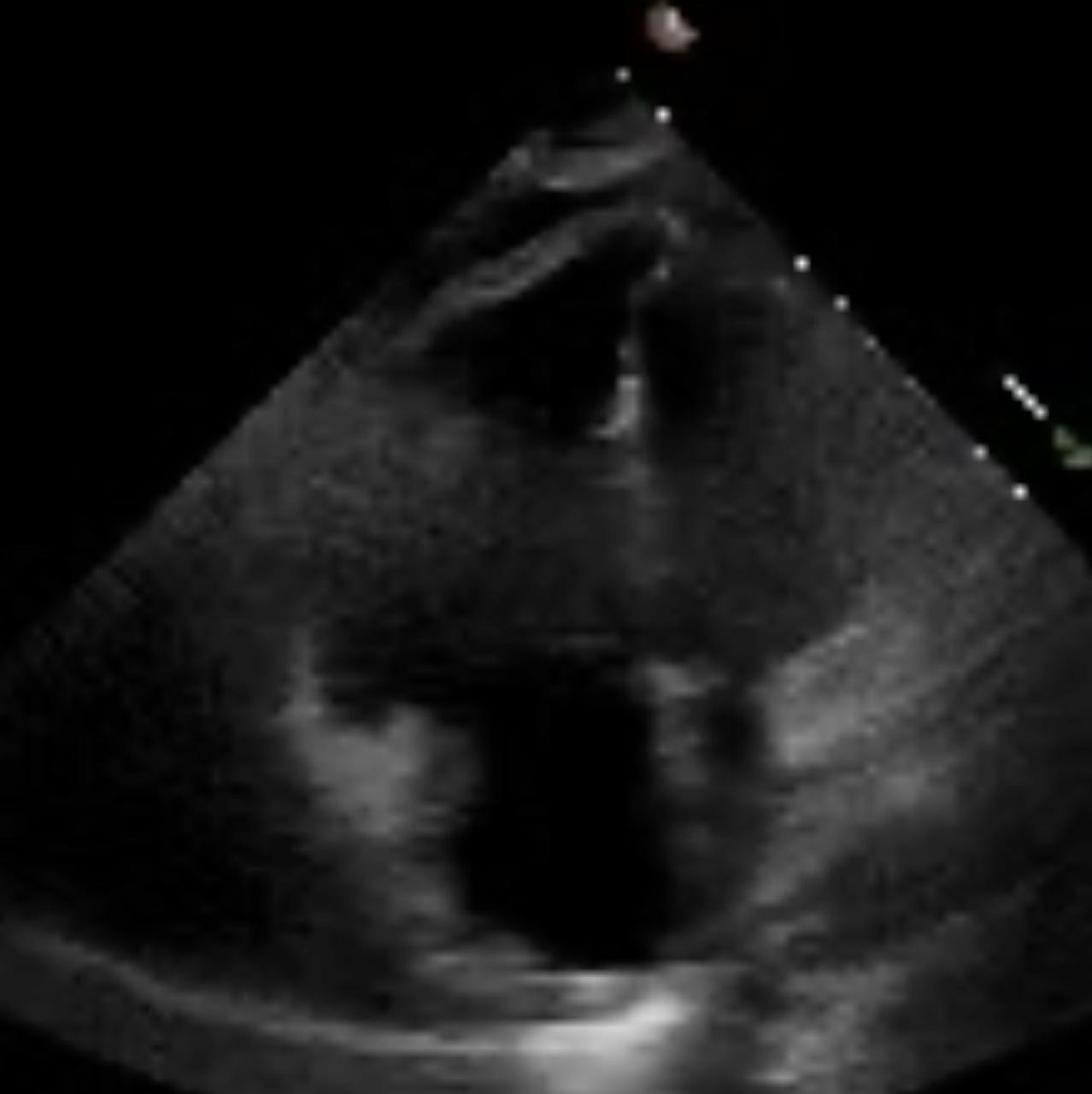}
    \includegraphics[width=0.3\linewidth]{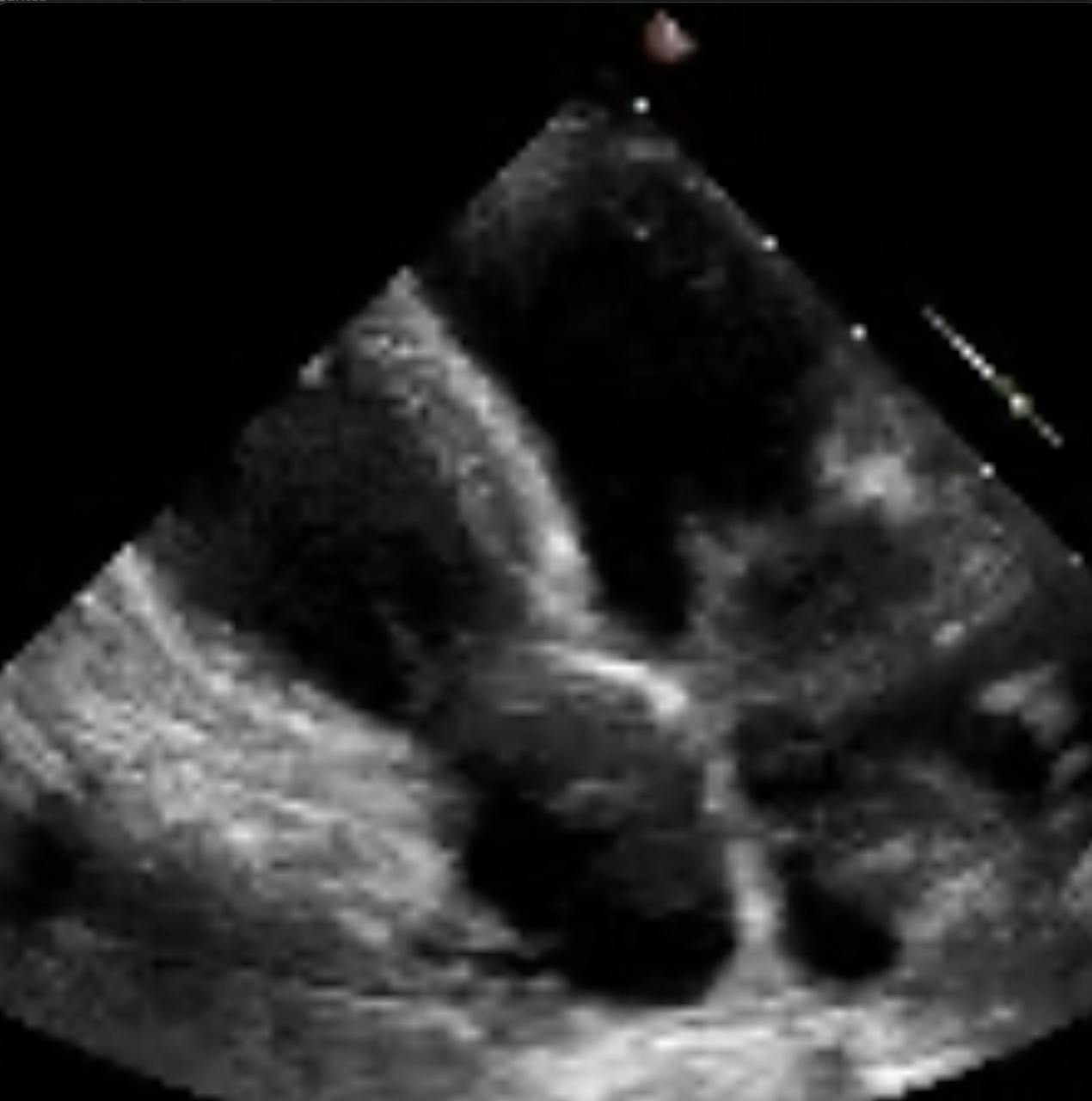}
    \includegraphics[width=0.3\linewidth]{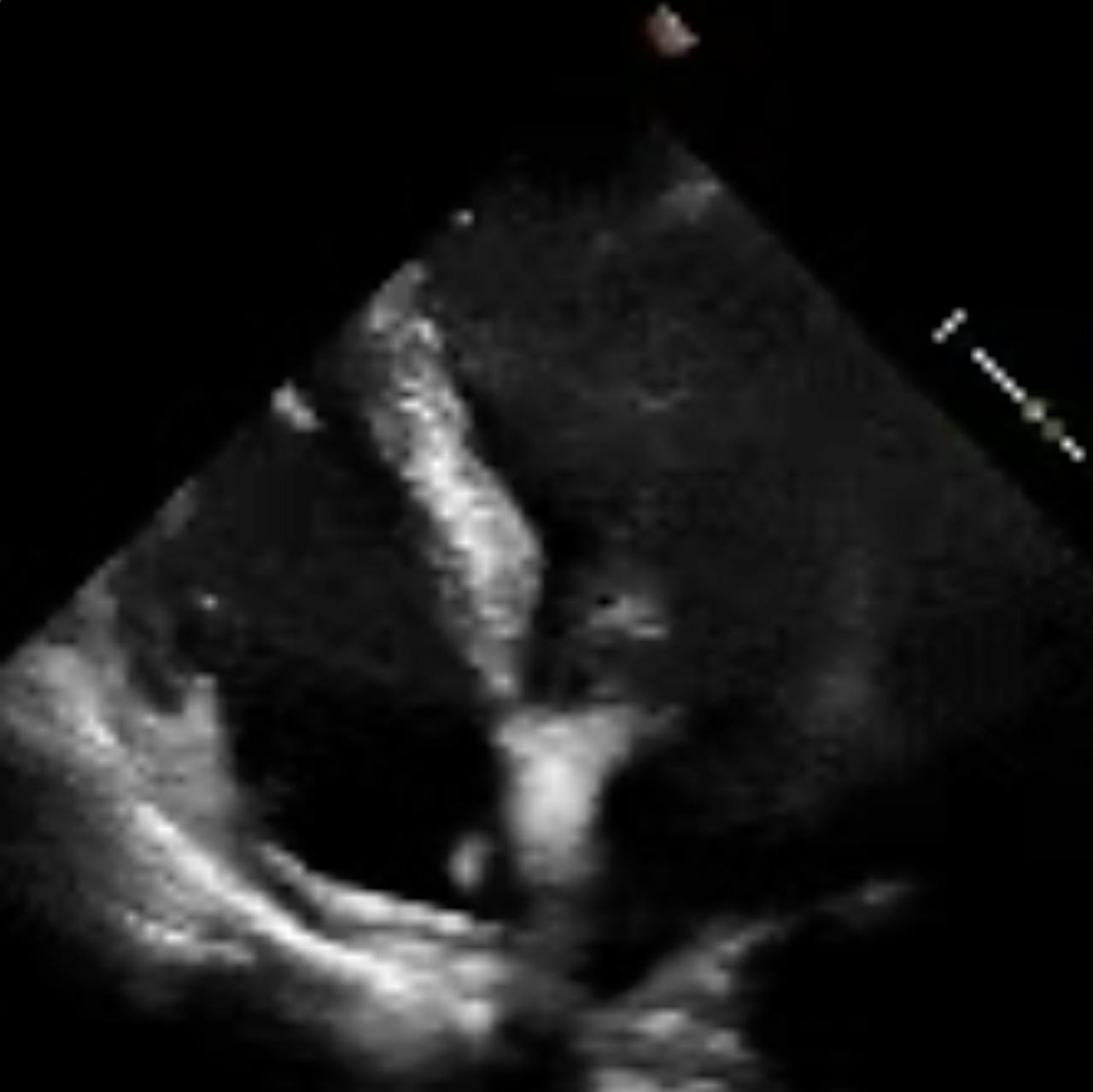} \\~\\
    Low ``foreshortening" response\\
    \includegraphics[width=0.3\linewidth]{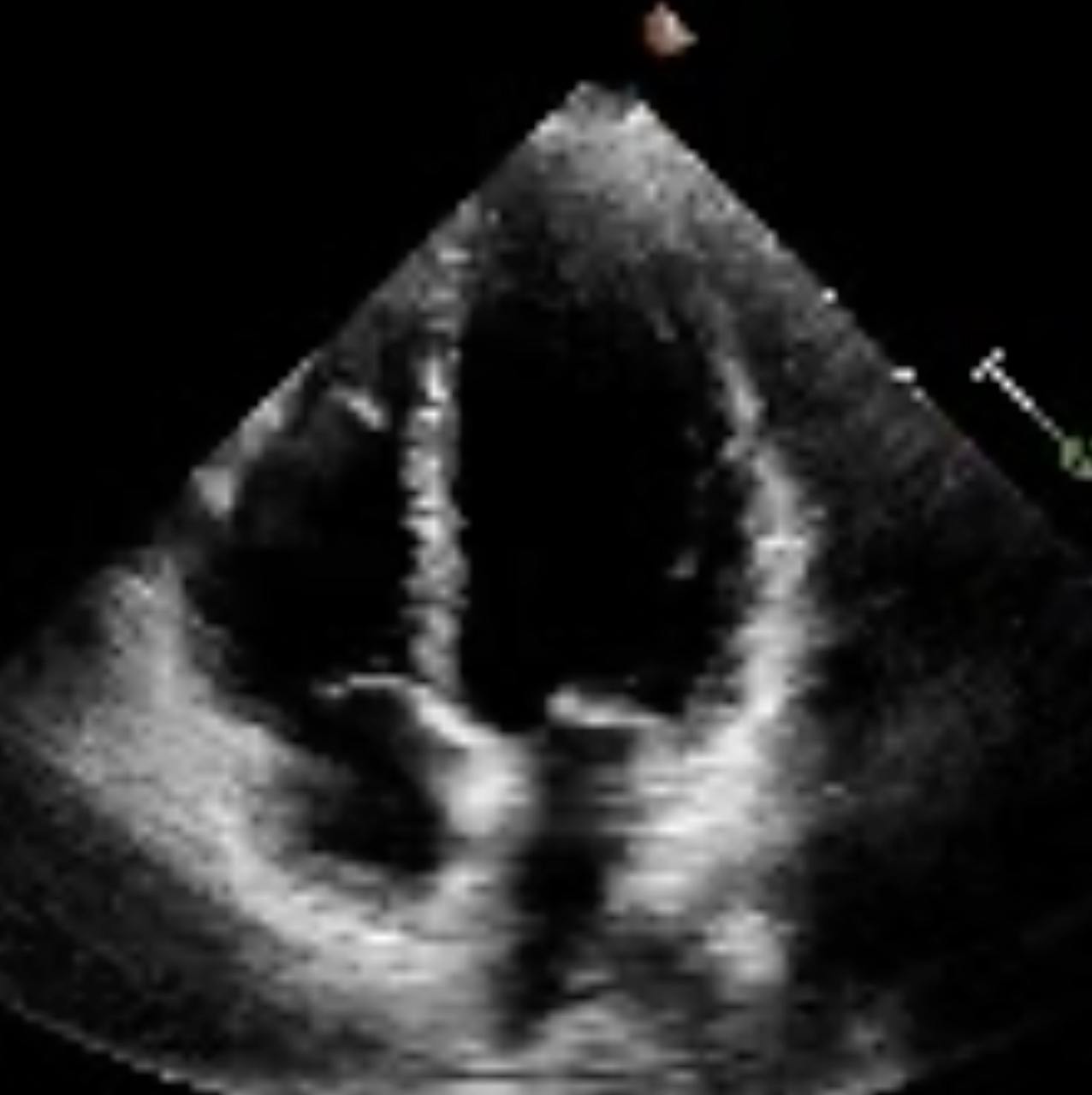}
    \includegraphics[width=0.3\linewidth]{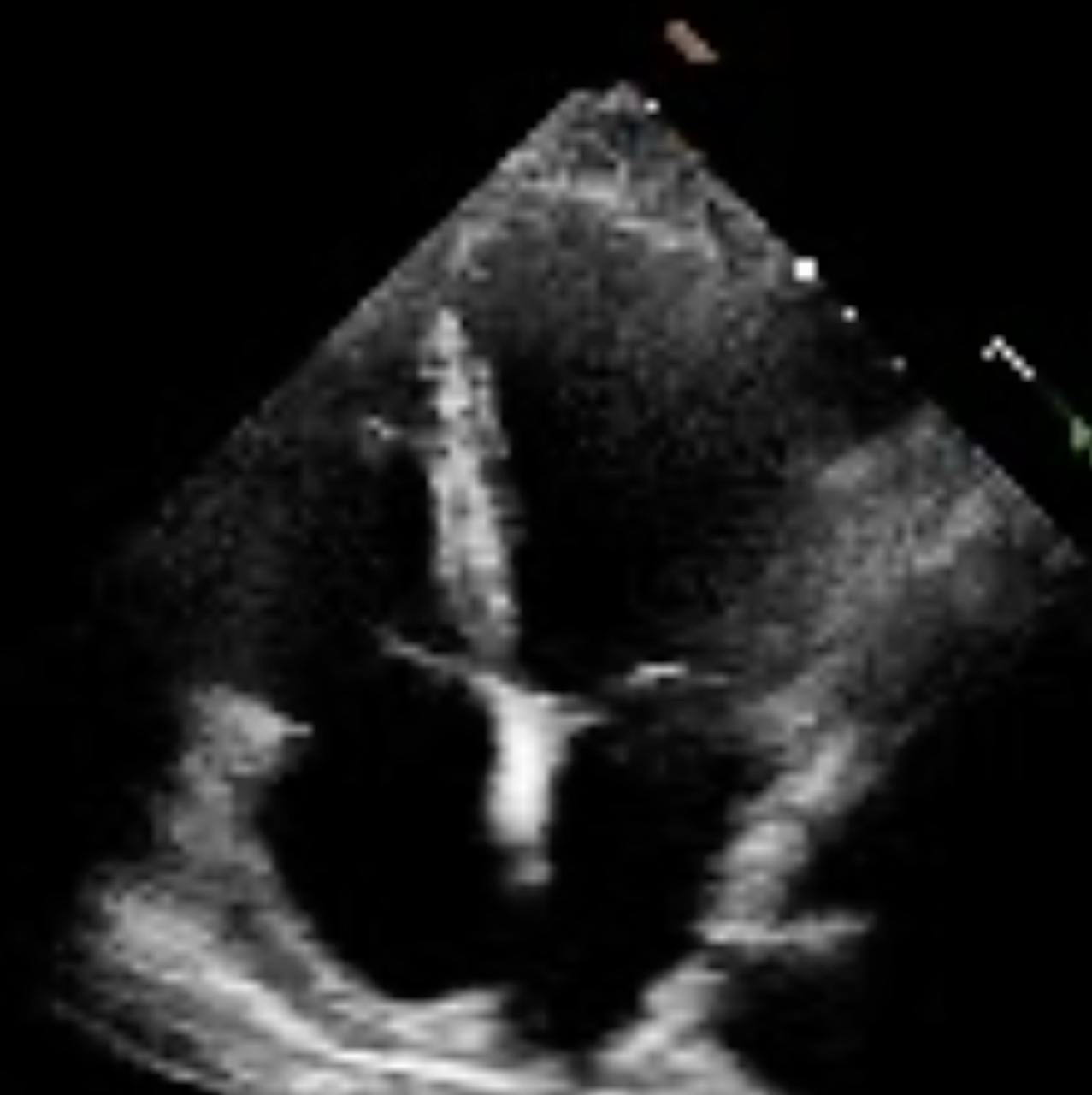}
    \includegraphics[width=0.3\linewidth]{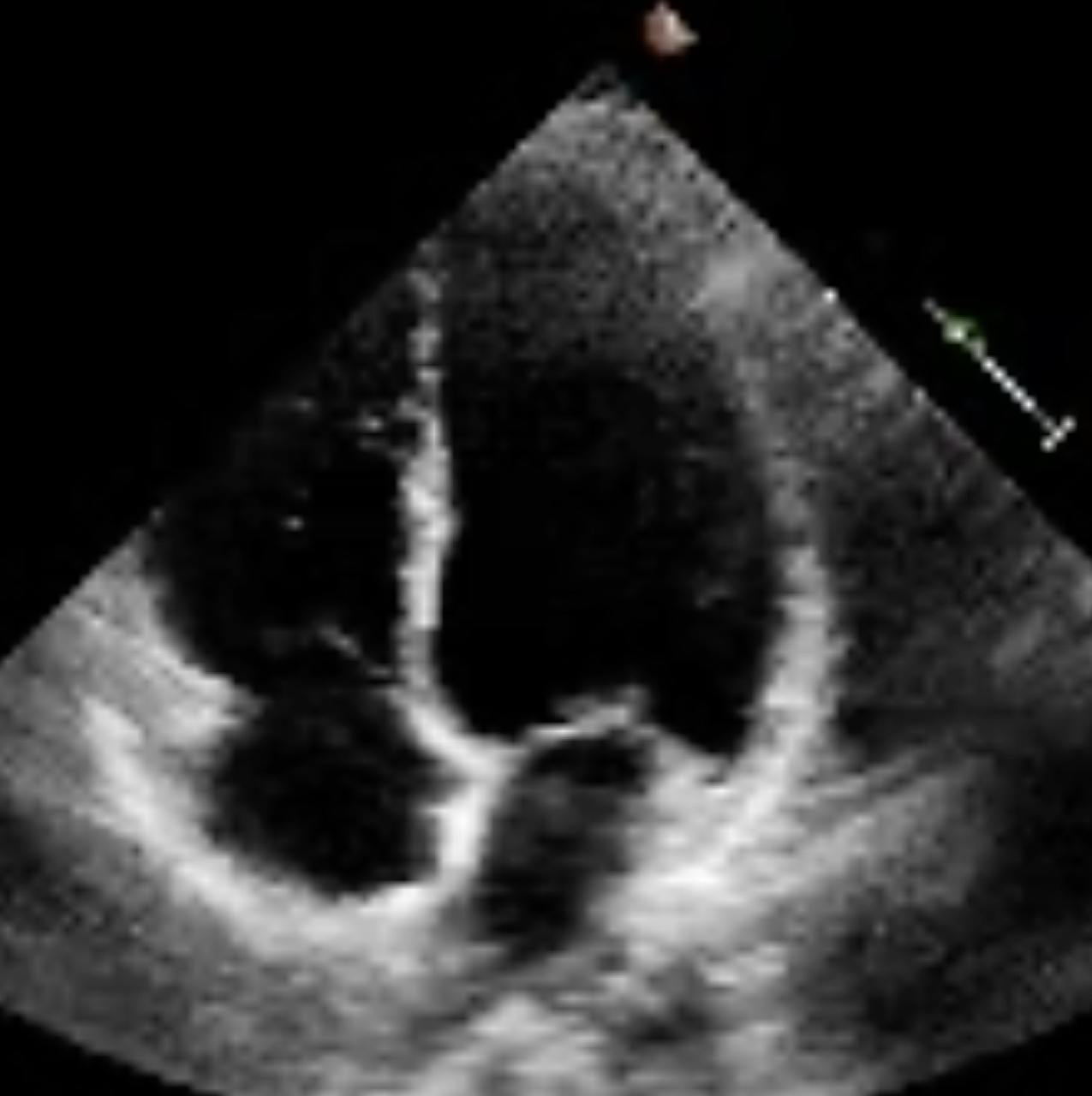}
    \caption{Example of samples depending on their EchoCLIP prompt response.}
    \label{fig:foreshortening}
\end{figure}

Our EchoCLIP~\cite{christensen2024echoclip} based raw results can be found under the \texttt{echoclip.csv} file of the \texttt{id636\_supplementary.zip} file. Figure~\ref{fig:foreshortening} represents the videos that activated the least and the most ``foreshortening" prompts.

\section{Portion of the dataset where the DFKD errors are located}
\label{sec:small_portion}
\begin{figure}
    \centering
    \resizebox{\linewidth}{!}{    \begin{tikzpicture}
        \begin{axis}[
            xlabel={Proportion of Test Set},        
            ylabel={Cumulative Proportion of Errors},  
            grid=both,                             
            width=10cm, height=7cm,                
            title={Proportion of Errors by Percentage of Test Set},
            ]
            
            \addplot[
                blue,
                thick,                              
                mark=*,                             
                mark size=1.5pt,                    
            ] table[
                col sep=comma,                      
                x={Proportion},                     
                y={CumulativeError}                 
            ] {figures/supplementary/cumulative_errors/plot.csv};                           
            
        \end{axis}
    \end{tikzpicture}}
    \caption{Proportion of squared errors as a function of test set percentage.}
    \label{fig:fails}
\end{figure}
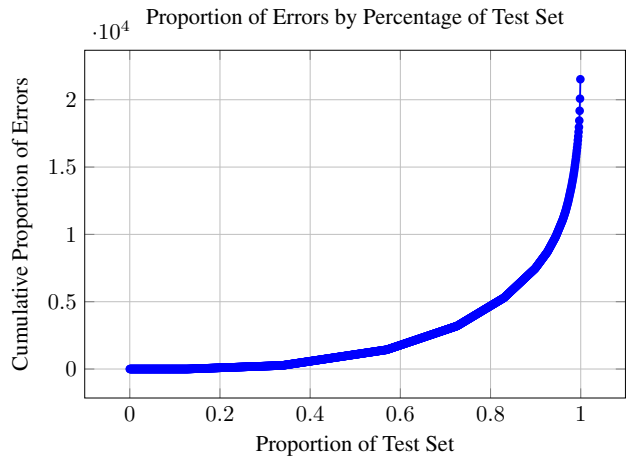

When studying the proportion of \ourmodel(DFKD) errors
Our analysis observed that the total error is concentrated within a small portion of the test set. As depicted in Figure~\ref{fig:fails}, the cumulative proportion of errors remains low for most of the test data, with a steep increase occurring in the final 15\% of the dataset. This indicates that \ourmodel~performs well across most of the test set, and the errors are predominantly localized to a specific subset.

\section{Scaling law}
\label{sec:scaling_law}
In Figures~\ref{fig:scaling_humans} and \ref{fig:scaling_echoclip}, we represented how 1-\meanIoU, 1-Dice score and \aFDED+\aFDES  scale with model weight. We observe that we reach a limit in performance around 1M parameters.

We obtain a log-slope of 0.15 for aFD versus human choice, 0.086 for aFD versus EchoCLIP, 0.16 for dice score, 0.11 for meanIoU. For comparison, in the regime with real data, aFD improves with a log-slope of 0.124 with human as a reference, 0.07 with EchoCLIP reference, 0.067 in meanIoU, and 0.088 in dice score. The slopes could provide insight into the dimension of the Riemannian manifold created by the model to handle its task. Still, it might be necessary to focus on precisely characterizing the boundary between the linear and saturation regimes and determining the theoretical limit with great precision to shift the logarithm instead of starting at 0 or 100\%. We can at least observe that the slopes are steeper when training on synthetic data. Since the slopes tends to be inversely proportional to the dimensions of the surfaces \cite{sharma2022scaling}, this is consistent with the idea that synthetic data tends to be less complex than real data.

\begin{figure}
    \centering
    \resizebox{\linewidth}{!}{\begin{tikzpicture}
    \begin{axis}[
        xlabel={$\log(\text{Number of parameters})$},  
        ylabel={$\log(\text{aFD}_{ED}+\text{aFD}_{ES})$},                            
        grid=both,                                     
        legend pos=north east,                         
        ]

        \addplot[
            only marks,
            mark=*,
            blue,  
        ] table[
            col sep=comma,                             
            x expr={ln(\thisrow{x})},                  
            y expr={ln(\thisrow{y})}                   
        ] {figures/supplementary/scaling_laws/afd_human/plot.csv};
        
        \addplot[
            red, 
            thick, 
            domain=8.5:15.1,                               
            samples=100                                
        ] {-0.15243921266547528*x + 3.796120263924857};
        
        \legend{,$y = - 0.15 x + 3.80$}
        
    \end{axis}
\end{tikzpicture}}~\\~\\
    \resizebox{\linewidth}{!}{\begin{tikzpicture}
    \begin{axis}[
        xlabel={$\log(\text{Number of parameters})$},  
        ylabel={$\log(1 - \text{Dice score})$},                            
        grid=both,                                     
        legend pos=north east,                         
        ]

        \addplot[
            only marks,
            mark=*,
            blue,  
        ] table[
            col sep=comma,                             
            x expr={ln(\thisrow{x})},                  
            y expr={ln(\thisrow{y})}                   
        ] {figures/supplementary/scaling_laws/dice_human/plot.csv};
        
        \addplot[
            red, 
            thick, 
            domain=8.5:15.1,                               
            samples=100                                
        ] {-0.15712181283569526*x + -0.23752531358050732};
        
        \legend{,$y = - 0.16 x -0.24$}
        
    \end{axis}
\end{tikzpicture}}~\\~\\
    \resizebox{\linewidth}{!}{\begin{tikzpicture}
    \begin{axis}[
        xlabel={$\log(\text{Number of parameters})$},  
        ylabel={$\log(1 - \text{meanIoU})$},                            
        grid=both,                                     
        legend pos=north east,                         
        ]

        \addplot[
            only marks,
            mark=*,
            blue,  
        ] table[
            col sep=comma,                             
            x expr={ln(\thisrow{x})},                  
            y expr={ln(\thisrow{y})}                   
        ] {figures/supplementary/scaling_laws/meanIoUs_human/plot.csv};
        
        \addplot[
            red, 
            thick, 
            domain=8.5:15.1,                               
            samples=100                                
        ] {-0.10506225977482185*x + -0.37638093327245103};
        
        \legend{,$y = - 0.11 x -0.38$}
        
    \end{axis}
\end{tikzpicture}}
    \caption{Scaling laws with humans as annotators.~\\}
    \label{fig:scaling_humans}
\end{figure}
\begin{figure}
    \centering
    \resizebox{\linewidth}{!}{\begin{tikzpicture}
    \begin{axis}[
        xlabel={$\log(\text{Number of parameters})$},  
        ylabel={$\log(\text{aFD}_{ED}+\text{aFD}_{ES})$},                            
        grid=both,                                     
        legend pos=north east,                         
        ]

        \addplot[
            only marks,
            mark=*,
            blue,  
        ] table[
            col sep=comma,                             
            x expr={ln(\thisrow{x})},                  
            y expr={ln(\thisrow{y})}                   
        ] {figures/supplementary/scaling_laws/afd_echo/plot.csv};
        
        \addplot[
            red, 
            thick, 
            domain=8.5:15.1,                               
            samples=100                                
        ] {-0.08630848076191498*x + 3.677995496872927};
        
        \legend{,$y = - 0.086 x + 3.68 $}
        
    \end{axis}
\end{tikzpicture}}
    \caption{Scaling laws with EchoCLIP as annotator.}
    \label{fig:scaling_echoclip}
\end{figure}
\section{First frames convergence}
\label{sec:first_frames}
\begin{figure}
 \centering
 Slow initial convergence\\\includegraphics[width=0.3\linewidth]{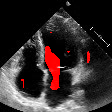}
 \includegraphics[width=0.3\linewidth]{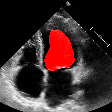}
 \includegraphics[width=0.3\linewidth]{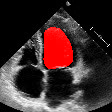} \\
 ~\\
 Very slow convergence\\\includegraphics[width=0.3\linewidth]{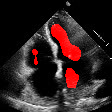}\includegraphics[width=0.3\linewidth]{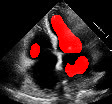}\includegraphics[width=0.3\linewidth]{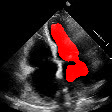}
 \includegraphics[width=0.3\linewidth]{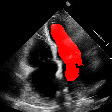}\includegraphics[width=0.3\linewidth]{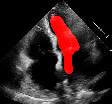}\includegraphics[width=0.3\linewidth]{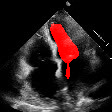}
 \includegraphics[width=0.3\linewidth]{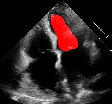}
 \includegraphics[width=0.3\linewidth]{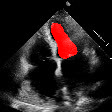}
 \caption{2 examples of slow convergence.}
 \label{fig:warmup}
\end{figure}

We mentioned in Section~4 of the main paper that one of the limitations of \ourmodel~was the few first frames for the model to converge to the solution, as depicted in Figure~\ref{fig:warmup}. In most cases, we can encounter that problem by taking the most significant connected component or prepadding the sequence to make the inferences converge.

\section{Multi-teacher}
\label{sec:multi}

Being able to accumulate multiple teachers to form a cohort opens up several possibilities, upon which the opportunity to refine target masks through ensemble learning and the ability to extend the single-task framework to multi-task learning~\cite{caruana1997multitask}.

Ensemble learning may require substantial computational resources for the model selection phase~\cite{caruana2004ensemble} and algorithms more sophisticated than simple averaging for combining the masks. Numerous variations of the standard STAPLE~\cite{warfield2004simultaneous} algorithm have been adapted to address the specificities of a segmentation task.

\begin{figure}[h]
    \centering
    \resizebox{\linewidth}{!}{
    \begin{subfigure}{0.15\textwidth}
        \includegraphics[width=\linewidth]{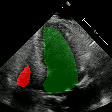}
        \caption*{(a)}
    \end{subfigure}
    \hfill
    \begin{subfigure}{0.15\textwidth}
        \includegraphics[width=\linewidth]{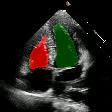}
        \caption*{(b)}
    \end{subfigure}
    \hfill
    \begin{subfigure}{0.15\textwidth}
        \includegraphics[width=\linewidth]{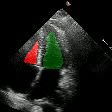}
        \caption*{(c)}
    \end{subfigure}
    }
    \caption{\ourmodel~outputs when trained to segment the two ventricles.}
    \label{fig:multi}
\end{figure}

Here, we focus on the potential of multi-task learning. We trained our model to replicate the left ventricle masks of DeepLabV3 from Echonet Dynamics (as in the rest of the paper) and the right ventricle masks from another model, EchoGAN. While EchoGAN can also generate left ventricle masks, it is less precise than DeepLabV3 trained on Echonet Dynamics, having been trained on ten times fewer examples. We achieved performance comparable to the main experiment for left ventricle segmentation while simultaneously providing our model with a basic capability in right ventricle segmentation.
For illustrative purposes, we show some outputs of the student model trained to segment the two ventricles in Figure~\ref{fig:multi}.

\section{Inference on CAMUS dataset}
\label{sec:camus}

The performance of \ourmodel~on CAMUS dataset is reported in Table~\ref{tab:camus}. Despite its small size and short sequences, which penalize our model's warm-up requirements, compared to SimLvSeg Dice score (0.906), EchoDFKD still performs well (0.852) even though it's trained on synthetic data with far fewer parameters.

\begin{table}[h!]
\centering
\resizebox{\linewidth}{!}{
\begin{tabular}{rrcccc}
\hline
  && B1 & B2 & B3 & B4 \\ \toprule
\multirow{4}{*}{meanIoU}    & l1 & 20.68\% & 68.89\% & 68.41\% & 73.37\% \\
                            & l2 & 53.64\% & 66.44\% & 70.70\% & 75.08\% \\
                            & l3 & 58.29\% & 64.46\% & 70.09\% & 72.69\% \\
                            & l4 & 63.63\% & 49.81\% & 72.66\% & 73.56\% \\ \midrule
\multirow{4}{*}{Dice score} & l1 & 29.50\% & 80.29\% & 79.95\% & 83.65\% \\
                            & l2 & 67.58\% & 78.21\% & 81.77\% & 85.21\% \\
                            & l3 & 72.23\% & 76.20\% & 81.55\% & 85.03\% \\
                            & l4 & 76.65\% & 62.08\% & 83.23\% & 84.17\% \\
\bottomrule
\end{tabular}
}
\caption{Traditional performance metrics across \ourmodel~configurations, on the CAMUS dataset.
}\label{tab:camus}
\end{table}

\end{document}